\documentclass[conference]{IEEEtran}

\usepackage{cite}
\usepackage{amsmath,amssymb}
\usepackage{graphicx}
\usepackage{textcomp}
\usepackage{xcolor}
\usepackage{booktabs}
\usepackage{colortbl}
\usepackage{makecell}
\usepackage{siunitx}
\usepackage{tabularray}
\UseTblrLibrary{booktabs}
\usepackage{listings}
\usepackage{url}
\usepackage{hyperref}
\usepackage{tikz}
\usetikzlibrary{positioning,calc,fit,decorations.pathreplacing,arrows.meta}
\usepackage{pgfplots}
\pgfplotsset{compat=1.18}
\usepackage[listings,skins,breakable]{tcolorbox}

\definecolor{figInputFill}{RGB}{218,213,204}
\definecolor{figAgentFill}{RGB}{178,194,218}
\definecolor{figOutputFill}{RGB}{198,213,228}
\definecolor{figEvalFill}{RGB}{212,208,222}
\definecolor{figBarFill}{RGB}{98,138,183}
\definecolor{figInputBorder}{RGB}{170,162,148}
\definecolor{figAgentBorder}{RGB}{110,135,170}
\definecolor{figOutputBorder}{RGB}{130,158,182}
\definecolor{figEvalBorder}{RGB}{148,140,168}

\definecolor{tblHeaderTint}{RGB}{220,228,240}
\definecolor{tblRowAlt}{RGB}{246,246,248}

\newcommand{\tblhead}{\rowcolor{tblHeaderTint}[\tabcolsep][\tabcolsep]}
\newcommand{\tblband}{\rowcolor{tblRowAlt}[\tabcolsep][\tabcolsep]}

\definecolor{codeFrame}{RGB}{90,115,150}
\definecolor{codeBack}{RGB}{245,247,250}
\definecolor{codeString}{RGB}{60,90,130}
\definecolor{codeKeyword}{RGB}{40,80,120}

\newtcblisting{promptbox}[1]{
  listing only,
  listing options={basicstyle=\ttfamily\scriptsize,
    breaklines=true, breakatwhitespace=false, breakindent=10pt,
    postbreak=\mbox{\hspace{0pt}$\hookrightarrow$\space},
    columns=fullflexible, keepspaces=true, showstringspaces=false},
  colback=codeBack,
  colframe=codeFrame,
  coltitle=white,
  colbacktitle=codeFrame,
  fonttitle=\sffamily\bfseries\small,
  title={#1},
  boxrule=0.5pt,
  arc=2pt,
  left=4pt, right=4pt, top=2pt, bottom=2pt,
  width=\linewidth,
  before=\par\medskip\noindent,
  after=\par\medskip,
  enhanced, breakable,
}

\newcommand{\numval}[2]{#2}
\newcommand{\lit}[1]{#1}

\newcommand{\finding}[2]{#2}
\newcommand{\modelid}[1]{\texttt{\expandafter\hyphenbreaks#1\relax}}
\def\hyphenbreaks#1{%
  \ifx#1\relax\else
    \if#1-\discretionary{-}{}{-}\else#1\fi
    \expandafter\hyphenbreaks
  \fi}

\begin{document}

\title{The Answer Path and the Grounding Instruction\\in LLM Question Answering over Knowledge Graphs}

\author{
\IEEEauthorblockN{Arquimedes Canedo}
\IEEEauthorblockA{Siemens Digital Industries Software\\
Princeton, NJ, USA\\
arquimedes.canedo@siemens.com}}

\maketitle

\begin{abstract}

A graph retrieval-augmented generation pipeline chooses which triples to put in the prompt, a syntax to write them in, an order to write them in, and a sentence telling the model what to do with them. We vary all four over six large language models and two knowledge-graph question answering benchmarks. Two of the four choices move the answer and the other two are flat.

The first is whether the answer path, the triples needed to reach the answer, is in the prompt at all. Holding the number of triples fixed and replacing every triple that is not on the chain with material from an unrelated entity changes answer accuracy by \numval{PrecDelta}{+0.003} F1, while removing the chain costs most of what the graph was worth. Retrieval budget belongs on recall, and precision in the range we can test buys nothing. There is no retriever in this study: subgraphs are constructed from gold SPARQL, so precision here names a property of the context we build rather than a setting on a system.

The second is the grounding instruction. With no facts in the prompt, telling a model to answer using only the provided facts drops F1 from \numval{PermNoCtxOverall}{0.299} to \numval{MatchedNoCtxOverall}{0.035}, a factor of \numval{SuppressionFactor}{8.63}. That figure describes an evaluation with an empty context arm rather than a working pipeline, and an experiment that applies the instruction to its context arm but not to its no-context baseline manufactures a spurious finding that graph context hurts at depth. We found one in our own results and retract it. Because a wrong subgraph goes undetected by the model, the write path into the knowledge base is an attack surface with the reach of prompt injection and none of its visible signature.

Syntax, triple order and subgraph size produce no effect we can measure at multi-hop depth. The comparison that would price the grounding instruction against correct context turns out not to be measurable with a format-sensitive answer scorer, because the instruction determines the response format, and we report it as an open contrast rather than a number.

\end{abstract}


\section{Introduction}
\label{sec:intro}

Graph retrieval-augmented generation (GraphRAG) pipelines retrieve subgraphs from a knowledge base, serialize them as text, and place them in the context window of a large language model (LLM)~\cite{edge2024graphrag}. A pipeline therefore makes four choices: which triples to include, how to serialize them, how to order them, and what the system prompt tells the model to do with them. This paper evaluates all four using the same corpus, models, and scorer to identify where practitioner effort has measurable value.

The results support investment in the first and last choices. The intermediate choices produce no measurable effect under the conditions tested.

\textbf{Terms.} \emph{Parametric knowledge} is the knowledge a model draws from its weights when no facts are supplied; \emph{parametric recall} is the act of drawing on that knowledge. An \emph{arm} is one condition in a comparison evaluated over the same questions as the other conditions. Most comparisons have two arms, while the subgraph-size experiment has three. In a \emph{no-context condition}, the facts block in an otherwise identical prompt is replaced by ``No facts are available,'' isolating the presence of graph context. \emph{Matched prompts} means that both conditions use the same system prompt, including the same grounding instruction.

The \emph{answer path} contains the triples traversed by a question's gold SPARQL query to reach its answer. We also refer to these as chain triples, matching the terminology used in our analysis code. A \emph{distractor triple} is a genuine triple from the question's entity neighbourhood that does not lie on the answer path. Answers are evaluated by token-level F1 against gold labels, defined as the harmonic mean of token precision and recall on a zero-to-one scale. Unless stated otherwise, intervals are ninety-five\% bootstrap intervals over questions.

Two prompt regimes recur throughout the study. The \emph{strict prompt}, common in research and production pipelines, tells the model to ``answer using only the provided facts'' and to quote the triples it used. The \emph{permissive prompt} removes that restriction and allows the model to combine the supplied information with its existing knowledge.

\textbf{Retrieval precision.} An inexpensive retriever commonly returns the answer chain together with unrelated material. Returning only relevant material is more costly and motivates reranking and threshold tuning. The precision experiment holds each prompt at \numval{SizeFiftyTarget}{50} triples, retains the chain in every condition, and replaces an increasing share of the remaining triples with material from an unrelated question's subgraph. Across the full sweep, answer F1 changes by only \numval{PrecDelta}{+0.003}. The null holds at every reasoning depth and among questions that the model could not answer without graph context. Removing the chain produces a large loss in accuracy. These results separate retrieval quality into two components with different consequences: preserving the answer path has substantial value, while improving the provenance of surrounding triples does not within the tested range.

\textbf{The grounding instruction.} With no facts available, the strict prompt scores F1~=~\numval{MatchedNoCtxOverall}{0.035}, compared with \numval{PermNoCtxOverall}{0.299} under the permissive prompt, for a difference of \numval{PromptSwingNoCtx}{0.264}. Models follow the restriction even when the prompt supplies no facts. This result characterizes an evaluation with an empty-context arm; it does not characterize a working pipeline whose retriever returns evidence. If the context arm receives the strict prompt while the no-context baseline receives a different prompt, the resulting contrast combines the effects of context and instruction while attributing both to the graph. An initial experiment built this way made context appear harmful at three and four hops, and that result is retracted here. When both arms use the strict prompt, context helps at one, two, and three hops. Its four-hop effect is non-negative but inconclusive.

\textbf{What is flat.} Once reasoning is required, serialization syntax, triple order, and subgraph size produce no measurable effects. Syntax appears to matter only for one-hop parsing, where the observed difference tracks whether the serializer provides a human-readable predicate name.

\textbf{What we cannot measure.} The remaining practical question is how much the strict instruction costs when the context is correct. The present apparatus cannot answer it. The strict prompt requires a structured evidence block, while the permissive prompt permits free-form answers. Prompt regime therefore determines response format, and the answer scorer is sensitive to that format. Changing the scorer's fallback reverses the sign of the contrast, with both estimates rejecting the null. We consequently report this contrast as unmeasurable with the current apparatus. Four results in earlier drafts depended on it.

Table~\ref{tab:variables} lists the variables varied by the experiments. The design makes them orthogonal, allowing each to change while the others remain fixed. Reasoning depth is the exception because it is a property of the question, not an experimental setting. In this corpus, depth is also perfectly aliased with the source benchmark.

\begin{table}[t]
\centering
\caption{Variables varied in the study and the corresponding summary findings. Depth is measured; all other variables are set experimentally.}
\label{tab:variables}
\begin{tabular}{llp{2.5cm}}
\toprule
\tblhead Variable & Levels & Effect \\
\midrule
\tblband Retrieval precision & \numval{PrecWrongPctZero}{0}--\numval{PrecWrongPctHundred}{100}\% wrong & \finding{RecallIsTheLever}{None at constant volume; recall is the lever} \\
Context & correct, wrong, none & \finding{ContextMattersWhenAnswerAbsent}{Large when the answer is absent. Suppression itself is small and model-dependent} \\
\tblband Prompt & strict, permissive & \finding{PromptIsLargestLever}{Largest lever with no context; not measurable with it} \\
Format & 7 serializations & \finding{FormatIsOneHopOnly}{Parsing only, aliased with labelling} \\
\tblband Triple order & 5 orders & None \\
Subgraph size & 0--200 triples & None \\
\tblband Model & 6, in 2 tiers & \finding{ModelModulatesFormat}{Modulates format} \\
Reasoning depth & 1--4+ hops & \finding{DepthGovernsTheRest}{Governs the rest, and is aliased with benchmark} \\
\bottomrule
\end{tabular}
\end{table}

The paper makes four contributions based on \numval{NumExperiments}{16} experiments comprising \numval{TotalTrials}{30,841} trials across six models from three vendors.

\begin{enumerate}
\item \textbf{\finding{SpendBudgetOnRecall}{For a retrieved subgraph, what matters is whether it contains the answer.}} With the chain present and the prompt fixed at \numval{SizeFiftyTarget}{50} triples, reducing context precision from \numval{PrecCtxPrecZero}{1.00} to \numval{PrecCtxPrecHundred}{0.51} changes answer F1 by \numval{PrecDelta}{+0.003} (95\% CI [\numval{PrecDeltaCILo}{\ensuremath{-}0.019}, \numval{PrecDeltaCIHi}{+0.031}], permutation $p$~=~\numval{PrecDeltaP}{0.812}). Removing the chain lowers F1 to \numval{OvPermWrong}{0.231} under the permissive prompt and \numval{OvStrictWrong}{0.005} under the strict prompt. For a pipeline operating under a fixed budget, these results support prioritizing recall. The conclusion also holds when the analysis is restricted to questions the model could not answer unaided.

\item \textbf{\finding{LargestLeverDependsOnContext}{The nine-fold grounding figure applies to the no-context condition and characterizes an evaluation, not a pipeline.}} The \numval{SuppressionFactor}{8.63}-fold suppression occurs in an empty-context arm. It is the largest effect measured in this study and is especially easy to misattribute to the graph. Any experiment that changes prompt regime together with context will assign the combined effect to context, as the retracted comparison did.

\item \textbf{\finding{OverrideIsModelDependent}{The raw comparison substantially overstates how much a wrong subgraph suppresses parametric recall.}} Replacing the correct subgraph with an unrelated one reduces F1 from \numval{OvPermCorrect}{0.627} to \numval{OvPermWrong}{0.231}, but almost all of this change reflects the loss of correct facts. Relative to a no-context baseline, which isolates suppression, the effect is \numval{OvVsNoCtx}{\ensuremath{-}0.068} (95\% CI [\numval{OvVsNoCtxCILo}{\ensuremath{-}0.098}, \numval{OvVsNoCtxCIHi}{\ensuremath{-}0.039}]). Moreover, \numval{OvModelsNoOverride}{2} of the six models show no suppression. \finding{PoisonedGraphIsWorseThanItLooks}{The performance that remains under a poisoned subgraph is parametric recall surviving the supplied context}; a graph the model has never encountered offers no such reserve.

\item \textbf{Four nulls and one unmeasurable contrast.} \finding{StopTuning}{Subgraph size, triple ordering, serialization and retrieval precision are not levers.} After matching question sets, size has no effect from chain-only to two hundred triples, and ordering has no effect at either tested scale. The serialization spread decreases from \numval{SpreadHopOne}{0.234} at one hop to \numval{SpreadHopThree}{0.036} at three hops. Separately, \finding{HandicapNotMeasurable}{the strict-versus-permissive contrast with correct context is not measurable with a format-sensitive answer scorer}, because the prompt regime also determines the response format.
\end{enumerate}

\begin{table*}[htbp]
\centering
\caption{Summary of the \numval{NumExperiments}{16} experiments, totaling \numval{TotalTrials}{30,841} trials across six models and three vendors. The $n$ column gives each experiment's denominator; the denominators overlap because several arms reuse Phase 1 as their control.}
\label{tab:summary}
\begin{tabular}{clllrl}
\toprule
\tblhead \# & Experiment & Lever Tested & Key Result & $n$ & Verdict \\
\midrule
\tblband 1 & Precision sweep & Wrong-provenance triples & $\Delta$=\numval{PrecDelta}{+0.003} at \numval{PrecCtxPrecHundred}{0.51} precision & \numval{PrecTrials}{1,999} & \finding{RecallIsTheLever}{Precision is not a lever} \\
2 & Wrong context (strict) & Context attribution & wrong=\numval{OvStrictWrong}{0.005}, correct=\numval{OvStrictCorrect}{0.579} & \numval{OverrideStrictTrials}{2,250} & \finding{WrongContextFallsToFloor}{Falls to no-context floor} \\
\tblband 3 & Wrong context (permissive) & Override, no strict instr. & wrong=\numval{OvPermWrong}{0.231}, correct=\numval{OvPermCorrect}{0.627} & \numval{OverridePermTrials}{1,500} & \finding{OverrideIsModelDependent}{Override model-dependent} \\
4 & Baseline (permissive) & No-context, permissive & F1=\numval{PermNoCtxOverall}{0.299} & \numval{NoCtxTrials}{750} & \finding{PermissiveBaselineUnfair}{Unfair comparison (retracted)} \\
\tblband 5 & Matched-prompt baseline & No-context, matched & 1-hop \numval{MatchedDeltaHopOne}{+0.809}, 4-hop \numval{MatchedDeltaHopFour}{+0.147} & \numval{MatchedTrials}{750} & \finding{ContextHelpsToThreeHop}{Helps 1--3 hop; 4-hop n.s.} \\
6 & Second template & Rephrased strict prompt & No-ctx F1=\numval{TemplateBNoCtx}{0.019} & \numval{TemplateBTrials}{1,500} & \finding{SuppressionConfirmed}{Suppression confirmed} \\
\tblband 7 & Format comparison & Serialization format & 1-hop spread \numval{SpreadHopOne}{0.234}, 3-hop spread \numval{SpreadHopThree}{0.036} & \numval{PhaseOneTrials}{5,250} & \finding{FormatIsOneHopOnly}{1-hop only, labelling aliased} \\
8 & Ordering comparison & Triple order & $\Delta$=\numval{LinDelta}{+0.006} & \numval{LinTrials}{3,000} & \finding{OrderingHasNoEffect}{No effect} \\
\tblband 9 & Triple order $\times$ size & Ordering at 200 triples & $\Delta_{200}$=\numval{LinDeltaSizeTwoHundred}{\ensuremath{-}0.005} & \numval{LinSizeTrials}{3,000} & \finding{OrderingNullAtScale}{No effect at scale} \\
10 & Distractor density & Subgraph size & Matched: \numval{SizeChainMatched}{0.752} / \numval{SizeTwentyFiveMatched}{0.737} / \numval{SizeFiftyMatched}{0.730}, all n.s. & \numval{SizeTrials}{1,842} & \finding{SizeIsSelectionArtifact}{No effect (selection artifact)} \\
\tblband 11 & Evidence-first & Prompt order & Faithfulness \numval{EvidenceFirstFaith}{+0.028}, F1 \numval{EvidenceFirstFI}{\ensuremath{-}0.021} & \numval{EvidenceFirstTrials}{1,500} & \finding{GroundingIsNotReasoning}{Grounding $\neq$ reasoning} \\
12 & Chain-of-thought & Step-by-step prompt & 2-hop \numval{CotDeltaHopTwo}{\ensuremath{-}0.020}, 3-hop \numval{CotDeltaHopThree}{\ensuremath{-}0.043} & \numval{CotTrials}{1,500} & \finding{CotHurts}{Hurts at all depths} \\
\tblband 13 & Dynamic oracle & Per-hop scoping & $\Delta$=\numval{DynDelta}{\ensuremath{-}0.057} vs full-context & \numval{DynamicTrials}{2,250} & \finding{ScopingLosesHolisticView}{Worse: loses holistic view} \\
14 & Temperature 0.7 & Sampling verification & Rankings preserved & \numval{TempTrials}{3,750} & \finding{GreedyResultsHold}{Greedy results hold} \\
\tblband 15 & SPARQL classification & Chain vs.\ fan-out & All 4 types present & \numval{NumQClassified}{43}q & \finding{StructureVariesWithinDepth}{Structure varies within depth} \\
16 & Instruction self-handicap & Strict vs.\ permissive & Arms differ in response format & \numval{HandicapTrials}{1,500} & \finding{HandicapNotMeasurable}{Not measurable with this scorer} \\
\bottomrule
\end{tabular}
\end{table*}

\section{Related Work}
\label{sec:related}

\subsection{Retrieval Quality in Graph Question Answering}

Dai et al.~\cite{dai2024llms} report that multi-hop accuracy collapses beyond two hops and that irrelevant triples do not reduce accuracy. Our precision sweep provides a controlled test of the latter result by holding triple count fixed and changing only provenance, thereby separating relevance from context volume. Mavromatis and Karypis~\cite{mavromatis2024gnnrag} adopt a different design, using a graph neural network to prune the retrieved subgraph before it reaches the model. Our findings provide no evidence that this pruning improves answer accuracy, although it may still reduce token use.

\subsection{Graph Serialization for LLMs}

Fatemi et al.~\cite{fatemi2024talk} showed that graph encoding choice produces accuracy differences of \lit{4.8}--\lit{61.8}\% on synthetic structural tasks across \lit{9} encodings and five PaLM models. Their encodings change naming and framing, including adjacency lists and sentences such as ``X and Y are friends''; they do not compare formal Resource Description Framework (RDF) serialization syntax. Sui et al.~\cite{sui2024table} compared table formats including CSV, JSON, HTML, and Markdown for single-hop LLM comprehension and found format-dependent accuracy. Frey et al.~\cite{frey2023turtle} directly benchmarked LLM comprehension of RDF syntax. Perozzi et al.~\cite{perozzi2024letgraph} instead represent structured data through learned soft prompts rather than textual serialization.

This study extends those lines of work to real knowledge-graph question answering with seven RDF-native formats. Its depth-stratified analysis also distinguishes parsing effects from reasoning effects. As in Fatemi et al., however, naming and syntax vary together in our design, leaving the two factors aliased.

\subsection{Knowledge Conflicts and Context Override}

Wu et al.~\cite{wu2024clasheval} found that LLMs override parametric knowledge with retrieved text roughly \lit{60}\% of the time. Xie et al.~\cite{xie2024chameleon} reported memorization ratios above \lit{80}\% for popular entities. Zou et al.~\cite{zou2024poisonedrag} corrupt a text corpus to steer retrieval-augmented generation (RAG) answers, while Zhao et al.~\cite{zhao2025ragsafety} showed that adversarial knowledge-graph (KG) triples reduce RoG's F1 from \lit{70}\% to \lit{38}\%, a relative decline of \lit{46}\%. Prior work uses either natural-language passages or individual adversarial triples; our experiment substitutes complete structured subgraphs.

Our result is not directly comparable to the figure reported by Wu et al. Their measure is an override rate, defined as the proportion of items for which the model followed retrieved text. Ours is a change in F1. Relative to a no-context baseline, the measured suppression is \numval{OvVsNoCtx}{\ensuremath{-}0.068} F1, and \numval{OvModelsNoOverride}{2} of the six models show no suppression. This evidence does not support treating suppression as an intrinsic property of structured context.

\subsection{Controlling Reliance on Context}

A related line of work studies how strongly a model should rely on retrieved text relative to its parameters and treats that reliance as a control objective. Zhou et al.~\cite{zhou2023context} showed that instruction phrasing alone changes context-faithfulness. Huang et al.~\cite{huang2024trust} formulate the objective as situated faithfulness, in which context is trusted when warranted. Bi et al.\ pursue the same objective through alignment~\cite{bi2024context} and fine-grained inference-time control~\cite{bi2025params}; Huang et al.~\cite{huang2025parammute} suppress knowledge-critical feed-forward layers. This literature supports calibrated reliance on context as an interior quantity worth tuning.

Our contribution to that work is methodological and negative. The endpoint prompts in our comparison produce different response formats, so an end-task scorer with unequal sensitivity to those formats cannot estimate the effect of moving between the endpoints. Section~\ref{sec:unmeasurable} documents this measurement failure.

In a controlled clinical setting, Mandarapu and Kunkunuru~\cite{mandarapu2026grounding} report that knowledge-graph grounding helps only when the relevant knowledge is absent from training. Because our entities come from Wikidata and were almost certainly represented in pretraining, their result predicts a small context benefit in this setting. We measure a large benefit because the comparison also depends on the prompt: their baseline may use parametric knowledge, while ours is forbidden to do so.

\subsection{Multi-Hop Knowledge Graph Question Answering}

Sun et al.~\cite{sun2024tog} introduced Think-on-Graph, and Jiang et al.~\cite{jiang2023structgpt} introduced StructGPT. Both show that LLM-driven iterative graph exploration can succeed on multi-hop tasks by allowing the model to control which triples it receives. Shan and Luo~\cite{shan2026bounded} found that bounded path history outperforms full history in iterative knowledge graph question answering (KGQA). The dynamic-oracle experiment in Section~\ref{sec:nulls} extends this line by showing that even oracle-correct per-hop scoping degrades performance when the decomposition is imposed externally. Nguyen et al.~\cite{nguyen2024cot} evaluate chain-of-thought directly on multi-hop knowledge-graph reasoning instead of relying only on end-task accuracy, and our chain-of-thought result is consistent with the pessimistic interpretation of that work.

\subsection{Capability Equalizer Effects}

Canedo~\cite{canedo2026equalizer} showed that architecture-specification format acts as a capability equalizer for code-generation agents, with format effects tenfold larger for weaker models. We observe an analogous pattern, but only for parsing. The format F1 spread is \numval{SpreadRatio}{2.6}$\times$ larger for mid-tier models than for frontier models. The effect disappears at multi-hop depths and remains subject to the same alias between syntax and predicate vocabulary disclosed in Section~\ref{sec:nulls}.\section{Experimental Setup}
\label{sec:setup}

\begin{figure}[htbp]
\centering
\includegraphics[width=\columnwidth]{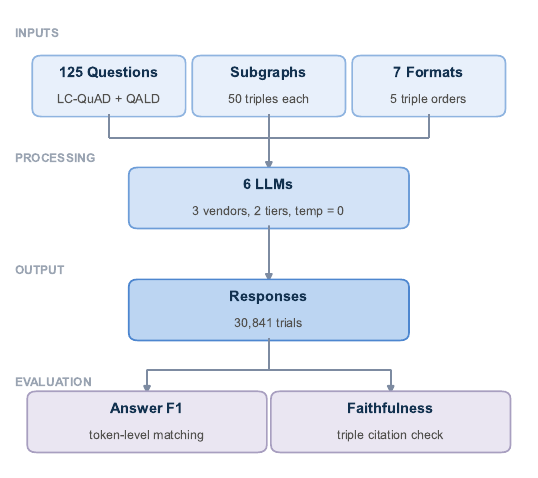}
\caption{Experimental pipeline. The \numval{NumQuestions}{125} questions have oracle subgraphs that are serialized in seven formats under five triple orders and sent to six LLMs. Responses are evaluated for answer F1 and evidence faithfulness.}
\label{fig:pipeline}
\end{figure}

Figure~\ref{fig:pipeline} summarizes the experimental pipeline.

\subsection{Questions and Subgraphs}

The corpus contains \numval{NumQuestions}{125} Wikidata-grounded questions~\cite{vrandecic2014wikidata} from two benchmarks: \numval{NumLcquad}{82} one- or two-hop questions from LC-QuAD 2.0~\cite{dubey2019lcquad2} and \numval{NumQald}{43} questions of at least three hops from the tenth edition of Question Answering over Linked Data (QALD)~\cite{usbeck2023qald10}. A gold SPARQL query identifies the reasoning chain for each question. We use it to extract an oracle subgraph of \numval{SizeFiftyTarget}{50} triples comprising the chain and distractors. The distractors are genuine Wikidata facts about nearby entities, so these subgraphs approximate plausible retriever output without using synthetic noise. The corpus includes \numval{NumQHopOne}{46} one-hop, \numval{NumQHopTwo}{36} two-hop, \numval{NumQHopThree}{29} three-hop, \numval{NumQHopFour}{12} four-hop, and \numval{NumQHopFiveSix}{2} five- or six-hop questions. The final \numval{NumQHopFiveSix}{2} questions remain in all aggregate results but are excluded from hop-stratified tables because that group is too small to interpret.

\textbf{\finding{DepthIsAliasedWithBenchmark}{Depth is completely aliased with benchmark in this corpus.}}
\label{sec:alias}
LC-QuAD supplies every one- and two-hop question, while QALD supplies every three- and four-hop question. Specifically, all \numval{NumQHopOne}{46} one-hop and \numval{NumQHopTwo}{36} two-hop questions come from LC-QuAD, and all \numval{NumQHopThree}{29} three-hop and \numval{NumQHopFour}{12} four-hop questions come from QALD. Consequently, no depth-stratified contrast in this paper can distinguish an effect of depth from a difference between the datasets. This qualification appears here because it applies to every depth-stratified result that follows.

The corpora also differ on several measured properties. LC-QuAD has \numval{LcGoldMean}{3.12} gold answers per question on average, compared with \numval{QaGoldMean}{7.70} for QALD. The answer is present in \numval{LcCovered}{76} of \numval{LcQuestions}{82} LC-QuAD subgraphs and in \numval{QaCovered}{16} of \numval{QaQuestions}{43} QALD subgraphs. With correct context, the models score \numval{LcCtxFone}{0.710} on LC-QuAD and \numval{QaCtxFone}{0.270} on QALD; without a graph, they score \numval{LcUnaided}{0.231} and \numval{QaUnaided}{0.427}, respectively. Thus, on QALD, unaided performance exceeds performance with the graph. Depth, answer-set size, coverage, difficulty, and the direction of the context effect all covary under this design.

Accordingly, a result reported ``at three-hop'' should be read as a result ``on QALD, at three-hop.'' We retain depth-stratified reporting because it matches the experimental design and conventions in prior work, but the evidence does not support recommending a depth threshold. Because depth-stratified knowledge-graph question answering corpora are commonly assembled by pooling benchmarks with different depth profiles, we expect this alias may also affect results beyond this study.

Two additional corpus properties qualify the depth-stratified analyses. Gold-answer coverage within the extracted subgraph is incomplete and declines with depth, while hop count measures predicate count and not a verified sequential chain. Section~\S\ref{sec:limitations} quantifies both properties.

\subsection{Models}

The study uses six models from three vendors and two capability tiers: Claude Sonnet 4.6 and Haiku 4.5 from Anthropic, GPT-5 and GPT-5 Mini from OpenAI, and Gemini 2.5 Pro and Gemini 2.5 Flash from Google. All calls use temperature~=~0 through a unified API gateway serving the three providers.

\subsection{Serialization Formats and Triple Ordering}

Each subgraph is represented in seven formats ranging from human-readable text to machine-oriented syntax: prose as natural-language sentences, tabular as pipe-delimited rows, Turtle~\cite{w3c2014turtle}, N-Triples~\cite{w3c2014ntriples}, Cypher, JSON-LD~\cite{w3c2014jsonld}, and RDF/XML~\cite{w3c2004rdfxml}. The selection reflects formats practitioners are likely to have available, not a systematic sampling of a serialization design space. Four are standard RDF serializations emitted directly by W3C-compatible toolchains. Cypher represents the property-graph ecosystem outside the RDF stack, while prose and pipe-delimited tables are common choices when GraphRAG implementations provide their own serializers. Table~\ref{tab:format_examples} presents the same three triples in each format.

Serialization determines syntax but not necessarily sequence. For each format, we test five orders: entity-centric grouping by subject, path-centric placement of chain triples first, breadth-first search (BFS) from the question entity, a seeded random shuffle, and alphabetical lexicographic sorting. Prose, N-Triples, tabular, and Cypher preserve the supplied triple order. Turtle, JSON-LD, and RDF/XML regroup triples by subject, partially overriding the requested sequence. Prose therefore provides the cleanest test of ordering effects.

\begin{table*}[htbp]
\centering
\caption{Seven representations of the same knowledge, illustrated with three triples about Sungkyunkwan University. Mean length per fifty-triple subgraph ranges from \numval{TokensProse}{644} tokens for prose to \numval{TokensNtriples}{2,383} for N-Triples; RDF/XML uses \numval{TokensRdfxml}{1,508}, and JSON-LD uses \numval{TokensJsonld}{1,642}. Length does not predict accuracy because N-Triples is both the longest representation and a top-tier format. Prose, tabular, Turtle, N-Triples, and Cypher include predicate names, whereas JSON-LD and RDF/XML expose only P-numbers.}
\label{tab:format_examples}
\begin{tblr}{
  colspec = {l X},
  width = \textwidth,
  rowsep = 3pt,
  row{1} = {font=\bfseries, bg=tblHeaderTint},
  row{3,5,7} = {bg=tblRowAlt},
  column{1} = {wd=1.8cm},
  column{2} = {font=\small\ttfamily},
  hline{1,Z} = {0.8pt},
  hline{2} = {0.5pt},
}
{Format} & {\textnormal{\rmfamily Serialized Output (3 of 50 triples)}} \\
{Prose} & {\textnormal{\rmfamily Sungkyunkwan University country South Korea. Sungkyunkwan University located in the administrative territorial entity Seoul. Sungkyunkwan University location Suwon.}} \\
{Tabular} & {Subject | Predicate | Object \newline Sungkyunkwan University | country | South Korea \newline Sungkyunkwan University | located in admin. territorial entity | Seoul \newline Sungkyunkwan University | location | Suwon} \\
{Turtle} & {wd:Q41085 rdfs:label "Sungkyunkwan University" ; \newline \hspace*{1em}wdt:P17 wd:Q884 ; \# South Korea \newline \hspace*{1em}wdt:P131 wd:Q8684 ; \# Seoul \newline \hspace*{1em}wdt:P276 wd:Q20714 . \# Suwon} \\
{N-Triples} & {\# Sungkyunkwan University -- country --> South Korea \newline <http://.../Q41085> <http://.../P17> <http://.../Q884> . \newline \# Sungkyunkwan University -- location --> Seoul \newline <http://.../Q41085> <http://.../P131> <http://.../Q8684> .} \\
{Cypher} & {CREATE (n0:Entity \{name: 'Sungkyunkwan University'\}) \newline CREATE (n1:Entity \{name: 'South Korea'\}) \newline CREATE (n0)-[:COUNTRY]->(n1) \newline CREATE (n0)-[:LOCATED\_IN]->(n2)} \\
{JSON-LD} & {\{"@id": "wd:Q41085", "rdfs:label": "Sungkyunkwan University", \newline \hspace*{1em}"wdt:P17": \{"@id": "wd:Q884", "rdfs:label": "South Korea"\}, \newline \hspace*{1em}"wdt:P131": \{"@id": "wd:Q8684", "rdfs:label": "Seoul"\}\}} \\
{RDF/XML} & {<rdf:Description rdf:about="wd:Q41085"> \newline \hspace*{1em}<rdfs:label>Sungkyunkwan University</rdfs:label> \newline \hspace*{1em}<wdt:P17 rdf:resource="wd:Q884"/> <!-- South Korea --> \newline \hspace*{1em}<wdt:P131 rdf:resource="wd:Q8684"/> <!-- Seoul --> \newline </rdf:Description>} \\
\end{tblr}
\end{table*}

\subsection{Prompt Templates}

\begin{promptbox}{Strict Prompt (Phase 1 and most experiments)}
System: You are a knowledge graph question
answering system. You will be given a set of
facts from a knowledge graph and a question.
Answer the question using only the provided
facts.

Respond in exactly this format:

Answer: <your answer>

Evidence:
- <subject> -- <predicate> --> <object>

Rules:
- The Answer must be a short phrase or
  comma-separated list.
- Each Evidence line must be a fact from the
  provided knowledge that you used.
\end{promptbox}

\begin{promptbox}{Permissive Prompt (override experiment)}
System: You are a knowledge graph question
answering system. You will be given some
relevant background information from a
knowledge graph and a question. Use the
provided information along with your own
knowledge to answer the question.
\end{promptbox}

These boxes reproduce the prompts exactly. They differ along two dimensions: the strict prompt confines the answer to the supplied facts and requires a specific response format. Section~\ref{sec:unmeasurable} examines how the formatting difference interacts with the scorer.

\subsection{Two-Axis Evaluation}

Each response receives separate scores for answer quality and evidence faithfulness.

\emph{Answer F1} compares the parsed response with the gold labels through token-level matching after lowercasing, article removal, date normalization, and alias resolution using a curated table. Multi-answer questions use optimal bipartite matching so that answer order does not affect the score.

\emph{Evidence faithfulness} is the fraction of cited triples that match triples in the supplied subgraph. The evaluator recognizes six citation forms---arrow, dash-arrow, colon, parenthetical, comma-separated, and natural language---and fuzzy-matches them to subgraph triples. The two-axis evaluation distinguishes a correct answer supported by valid evidence, which is consistent with reading and reasoning over the graph; a correct answer paired with fabricated evidence, which is consistent with parametric recall followed by confabulated citations; and an incorrect answer supported by valid evidence, where relevant triples were found but not composed into the gold answer.

Answer parsing uses a primary path and a fallback. The primary path extracts a line beginning with ``Answer:'', as required by the strict prompt, while the fallback reads a free-form response. The prompt affects response format, and response format determines which parsing path runs. Section~\ref{sec:unmeasurable} describes the resulting measurement problem.

\section{Retrieval Precision Does Not Matter}
\label{sec:precision}

Retrieval can fail by omitting the answer chain or by returning that chain with additional material that is not on it. Although both failures are commonly summarized as retrieval quality and addressed with the same tuning controls, the results below show that they have different consequences.

\subsection{Design}

We fix each subgraph at \numval{SizeFiftyTarget}{50} triples and vary its composition while keeping the answer chain present. The other slots initially contain neighbours associated with the question. At each step of the sweep, a fraction of those triples is replaced with triples from another question's subgraph, using the deterministic donor mapping also used in the wrong-context experiments. The replaced fraction is \numval{PrecWrongPctZero}{0}, \numval{PrecWrongPctTwentyFive}{25}, \numval{PrecWrongPctFifty}{50}, or \numval{PrecWrongPctHundred}{100}\% of the available slots. This lowers context precision from \numval{PrecCtxPrecZero}{1.00} to approximately \numval{PrecCtxPrecHundred}{0.51}.

Because every condition contains the same number of prompt triples, the sweep changes provenance without changing volume. Each condition includes every question and model, yielding \numval{PrecTrials}{1,999} trials across \numval{PrecModels}{4} models and \numval{NumQuestions}{125} questions, with prose serialization and entity-centric ordering.

The full-precision condition reproduces the standard correct-context condition through a second, independently implemented runner. Its F1 is within \numval{PrecReplicationGap}{0.002} of the Phase One prose average. Read that as reassurance rather than as a validation: the sweep covers \numval{PrecModels}{4} models and the Phase One average covers six, so the two figures are unpaired means over different model sets and the agreement could as easily be produced by the two absent models as by the runners agreeing. A validation would recompute Phase One on the same four models, paired by question, and we have not done it. Nothing in the sweep's own contrast depends on this, since both of its arms come from the same runner.

All conditions use the strict prompt and therefore induce the same response format. The scorer issue described in Section~\ref{sec:unmeasurable} does not affect this comparison, and disabling the fallback parser leaves the results unchanged.

\subsection{Result}

Answer accuracy does not measurably change across the sweep. Replacing every distractor slot with triples from an unrelated entity neighbourhood changes answer F1 by \numval{PrecDelta}{+0.003} (95\% CI [\numval{PrecDeltaCILo}{\ensuremath{-}0.019}, \numval{PrecDeltaCIHi}{+0.031}], permutation $p$~=~\numval{PrecDeltaP}{0.812}, \numval{PrecDeltaNq}{125} questions). Table~\ref{tab:precision} reports all conditions. The null also holds within each reported depth: \numval{PrecDeltaHopOne}{+0.030} at one hop, \numval{PrecDeltaHopTwo}{\ensuremath{-}0.021} at two hops, \numval{PrecDeltaHopThree}{\ensuremath{-}0.007} at three hops, and \numval{PrecDeltaHopFour}{\ensuremath{-}0.001} at four hops. Across the four models, individual deltas range from \numval{PrecDeltaWorst}{\ensuremath{-}0.025} to \numval{PrecDeltaBest}{+0.014}.

Two restricted analyses address cases that could make the pooled null uninformative.

The first restriction concerns answer coverage. Fixing the chain triples does not guarantee that the subgraph contains every answer: only \numval{CovFullTotal}{92} of \numval{CovAnswerable}{117} answerable questions have all gold answer labels in their subgraph. For questions whose context never contained the answer, changing the surrounding triples may have no answer to disrupt. Among the \numval{PrecCoveredNq}{92} fully covered questions, the delta remains \numval{PrecCovered}{+0.004} (95\% CI [\numval{PrecCoveredCILo}{\ensuremath{-}0.027}, \numval{PrecCoveredCIHi}{+0.037}], $p$~=~\numval{PrecCoveredP}{0.799}); it is \numval{PrecUncovered}{\ensuremath{-}0.001} for the \numval{PrecUncoveredNq}{33} questions without full coverage. The covered subset is not constrained by a floor, scoring \numval{PrecAnswerZero}{0.579} under full context precision.

The second restriction concerns parametric knowledge. If a model has memorized an answer, it may ignore degraded context, allowing parametric recall to conceal a genuine precision effect in the pooled result. The same null appears on the \numval{PrecNovNothingNq}{48} questions for which unaided performance was zero and the model therefore had to use the supplied information: the delta is \numval{PrecNovNothing}{+0.015} (95\% CI [\numval{PrecNovNothingCILo}{\ensuremath{-}0.035}, \numval{PrecNovNothingCIHi}{+0.066}]). This subset approximates the knowledge conditions that would apply throughout a private graph.

\begin{table}[t]
\centering
\caption{Retrieval precision at constant volume. Each row contains the same number of triples and differs only in the fraction imported from another question's neighbourhood. Answer accuracy remains flat, while evidence faithfulness declines slightly.}
\label{tab:precision}
\begin{tabular}{lccc}
\toprule
\tblhead Wrong triples & Context precision & Answer F1 & Faithfulness \\
\midrule
\tblband \numval{PrecWrongPctZero}{0}\%   & \numval{PrecCtxPrecZero}{1.00} & \numval{PrecAnswerZero}{0.579} & \numval{PrecFaithZero}{0.914} \\
\numval{PrecWrongPctTwentyFive}{25}\%  & \numval{PrecCtxPrecTwentyFive}{0.86} & \numval{PrecAnswerTwentyFive}{0.584} & \numval{PrecFaithTwentyFive}{0.902} \\
\tblband \numval{PrecWrongPctFifty}{50}\%  & \numval{PrecCtxPrecFifty}{0.73} & \numval{PrecAnswerFifty}{0.583} & \numval{PrecFaithFifty}{0.893} \\
\numval{PrecWrongPctHundred}{100}\% & \numval{PrecCtxPrecHundred}{0.51} & \numval{PrecAnswerHundred}{0.583} & \numval{PrecFaithHundred}{0.888} \\
\bottomrule
\end{tabular}
\end{table}

\finding{MeasureFaithfulnessSeparately}{Answer accuracy and evidence faithfulness are distinct axes that can move independently.} Faithfulness is the only measured outcome that declines across this sweep, falling from \numval{PrecFaithZero}{0.914} to \numval{PrecFaithHundred}{0.888} as context precision decreases. Models cite irrelevant triples somewhat more often while continuing to derive correct answers from the retained chain. An answer metric alone will not reveal this deterioration, so systems that expose evidence to users should evaluate faithfulness separately.

\subsection{Where the null stops}

Together with Section~\ref{sec:override}, this result separates two components of retrieval quality. Removing the answer chain is catastrophic: wrong-context F1 falls to \numval{OvPermWrong}{0.231} with a permissive prompt and \numval{OvStrictWrong}{0.005} with a strict prompt. Keeping the chain while filling the remaining prompt capacity with unrelated triples has no measured answer cost. For a pipeline operating under a fixed budget, the evidence therefore supports allocating that budget to retrieval recall.

This conclusion is bounded by the experimental construction. Every condition in the precision sweep contains the answer chain, so the result applies to retrievers that find the answer and return additional irrelevant material. It does not apply when retrieval omits the answer, which the wrong-context experiment shows is harmful. On this evidence, improving precision at the expense of recall offers no measured benefit and introduces substantial risk.

Two additional qualifications apply. The wrong triples for each question come from one donor question and are therefore more internally coherent than errors drawn from many unrelated sources; dispersed noise could behave differently. The maximum replacement fraction is also limited by chain length, which averages \numval{PrecChainMean}{23} of the \numval{SizeFiftyTarget}{50} triples. Consequently, the \numval{PrecWrongPctHundred}{100}\% condition retains the complete chain and contains wrong material in only roughly half of the subgraph, so it is not a wholly incorrect context.\section{The Grounding Instruction Suppresses Parametric Recall}
\label{sec:suppression}

The strict prompt limits answers to the supplied facts, and models follow that instruction even when no facts are supplied.

With the facts block replaced by ``No facts are available,'' models under the strict prompt score F1~=~\numval{MatchedNoCtxOverall}{0.035} across \numval{MatchedTrials}{750} trials and all six models. On the same questions, the permissive prompt yields \numval{PermNoCtxOverall}{0.299}. The difference is a factor of \numval{SuppressionFactor}{8.63} and a swing of \numval{PromptSwingNoCtx}{0.264} F1, the largest single effect measured in this study. The scorer's fallback path rescues no trial in either arm, so disabling it leaves the contrast unchanged. A rephrased strict template (``Base your answer exclusively on the facts provided below'') produces slightly stronger suppression: no-context F1 is \numval{TemplateBNoCtx}{0.019} over \numval{TemplateBTrials}{1,500} trials, compared with Template~A's \numval{MatchedNoCtxOverall}{0.035}. Context benefit remains comparable across the templates (\numval{TemplateBCtx}{0.560} against \numval{TemplateACtx}{0.581}, both on prose). \finding{SuppressionConfirmed}{The suppression therefore does not depend on the specific strict template.}

\subsection{What the figure describes}

\finding{LargestLeverDependsOnContext}{This effect is specific to the no-context condition.} It characterizes a model instructed to ground its answer in facts that are absent, a condition created by an evaluation but not by a working pipeline. It does not measure the cost of the instruction when a retriever supplies the correct subgraph. Section~\ref{sec:unmeasurable} examines that second contrast and concludes that the present apparatus cannot measure it.

The no-context result remains important because it can invalidate an evaluation whose arms use different prompts.

\subsection{The artifact}
\label{sec:artifact}

The initial comparison paired a correct subgraph under the strict prompt with a no-context baseline that omitted both the ``use only the provided facts'' restriction and the structured evidence format. In that comparison, context appeared beneficial at one and two hops but harmful at three hops ($\Delta$F1~=~\numval{ArtifactDeltaHopThree}{\ensuremath{-}0.082}) and four hops ($\Delta$F1~=~\numval{ArtifactDeltaHopFour}{\ensuremath{-}0.338}). The result was interpreted as evidence that text-serialized subgraphs interfere with multi-hop reasoning.

That comparison changed context and prompt regime simultaneously, and the prompt has the larger effect. \finding{PermissiveBaselineUnfair}{A permissive no-context baseline is not a fair comparator for a strict context arm.} The result does not establish worse reasoning over the graph. The strict arm prohibited fallback to parametric knowledge, while the baseline allowed it.

\finding{ContextHelpsToThreeHop}{When both arms use the strict prompt, context helps at one, two and three hops; at four hops the estimate is non-negative but inconclusive} (Table~\ref{tab:matched}). The benefit declines from $\Delta$F1~=~\numval{MatchedDeltaHopOne}{+0.809} at one hop to \numval{MatchedDeltaHopFour}{+0.147} at four hops without becoming negative. It is significant at one, two and three hops under an exact sign-flip permutation test, with $p$~\numval{MatchedPermPHopThree}{\ensuremath{<}0.001} in each case. At four hops, $p$~=~\numval{MatchedPermPHopFour}{0.109} across \numval{NumQHopFour}{12} questions. Of those questions, \numval{MatchedHopFourNoBenefitQ}{7} show no benefit, and dropping the two largest deltas reduces the mean from \numval{MatchedDeltaHopFour}{+0.147} to \numval{MatchedDeltaHopFourTrimmed}{+0.028}. Excluding the two models whose truncated trials were re-run gives \numval{MatchedDeltaHopFourNoOpenAI}{+0.096} at $p$~=~\numval{MatchedPermPHopFourNoOpenAI}{0.328}. The four-hop claim is limited to context not hurting; the evidence does not establish a measurable benefit. Although the bootstrap interval excludes zero, we do not rely on it because the percentile bootstrap is anti-conservative for a sample this small, with five exact zeros and two large positives.

The arms in Table~\ref{tab:matched} are not matched on format. The no-context arm uses prose only, while the correct-context arm averages seven serializations. Section~\ref{sec:nulls} estimates the format spread at three hops as \numval{SpreadHopThree}{0.036} F1, well within these deltas, so the format mismatch does not account for them.

\begin{table}[t]
\centering
\caption{Context benefit under matched prompts, both conditions strict. No-context is the matched baseline (\numval{MatchedTrials}{750} trials), correct-context is all of Phase 1 (\numval{PhaseOneTrials}{5,250} trials). All values are answer F1. 95\% bootstrap CI in brackets; $p$ is an exact sign-flip permutation test, which we rely on in preference to the interval at 4-hop.}
\label{tab:matched}
\begin{tabular}{lcccc}
\toprule
\tblhead Condition & 1-hop & 2-hop & 3-hop & 4-hop \\
\midrule
\tblband Matched no-context & \numval{MatchedNoCtxHopOne}{0.014} & \numval{MatchedNoCtxHopTwo}{0.046} & \numval{MatchedNoCtxHopThree}{0.044} & \numval{MatchedNoCtxHopFour}{0.061} \\
Correct context    & \numval{CtxHopOne}{0.823} & \numval{CtxHopTwo}{0.567} & \numval{CtxHopThree}{0.311} & \numval{CtxHopFour}{0.208} \\
\tblband $\Delta$           & \numval{MatchedDeltaHopOne}{+0.809} & \numval{MatchedDeltaHopTwo}{+0.521} & \numval{MatchedDeltaHopThree}{+0.268} & \numval{MatchedDeltaHopFour}{+0.147} \\
{\scriptsize 95\% CI} & {\scriptsize[\numval{MatchedDeltaHopOneCILo}{0.73},\numval{MatchedDeltaHopOneCIHi}{0.88}]} & {\scriptsize[\numval{MatchedDeltaHopTwoCILo}{0.39},\numval{MatchedDeltaHopTwoCIHi}{0.64}]} & {\scriptsize[\numval{MatchedDeltaHopThreeCILo}{0.14},\numval{MatchedDeltaHopThreeCIHi}{0.40}]} & {\scriptsize[\numval{MatchedDeltaHopFourCILo}{0.003},\numval{MatchedDeltaHopFourCIHi}{0.32}]} \\
\tblband {\scriptsize perm.\ $p$} & {\scriptsize \numval{MatchedPermPHopOne}{\ensuremath{<}0.001}} & {\scriptsize \numval{MatchedPermPHopTwo}{\ensuremath{<}0.001}} & {\scriptsize \numval{MatchedPermPHopThree}{\ensuremath{<}0.001}} & {\scriptsize \numval{MatchedPermPHopFour}{0.109}} \\
{\scriptsize $n$ questions} & {\scriptsize \numval{NumQHopOne}{46}} & {\scriptsize \numval{NumQHopTwo}{36}} & {\scriptsize \numval{NumQHopThree}{29}} & {\scriptsize \numval{NumQHopFour}{12}} \\
\bottomrule
\end{tabular}
\end{table}

\begin{figure}[htbp]
\centering
\includegraphics[width=\columnwidth]{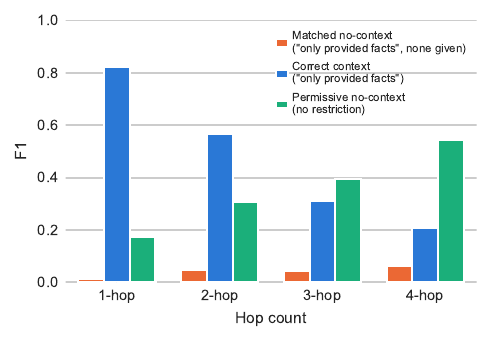}
\caption{Three baselines reveal the artifact. Strict prompt with no facts (F1~=~\numval{MatchedNoCtxHopOne}{0.014}--\numval{MatchedNoCtxHopFour}{0.061}); strict prompt with correct subgraph; and permissive prompt with no facts, which is unrestricted parametric recall. The original ``context hurts'' finding compared the strict-with-context series against the permissive-no-context one, conditions that differ in prompt regime and not just in context presence.}
\label{fig:instruction}
\end{figure}

Figure~\ref{fig:instruction} presents the three baselines that separate the two changes. The separation between strict context and permissive no context at three and four hops was initially attributed to context interference. It instead measures the difference between instructed and unrestricted parametric recall.

\subsection{Why the design is easy to build}

The design follows from two individually plausible choices. A context arm receives the strict prompt to ensure that the model reads the graph. A no-context arm is then created by removing the facts, often along with the instruction to use them because that instruction appears inapplicable once the facts are gone. Each arm is coherent in isolation, but together they vary prompt regime and context presence at the same time. At depths where parametric knowledge is strong, any experiment with this construction will therefore make context appear harmful.

We treat the prevalence of this design in other work as a hypothesis, not a finding. Establishing prevalence would require coding published RAG evaluations according to whether their no-context baselines retain the prompt used by their context conditions. We have not conducted that review. The issue warrants attention because a results table does not show which of the two changes produced the difference.

\subsection{The retraction is narrower than it first read}

The matched comparison above uses the strict prompt in both arms. Using a permissive prompt for the context arm produces a different pattern: \numval{PermCtxHopOne}{+0.699} at one hop and \numval{PermCtxHopTwo}{+0.234} at two hops, no measured effect at three hops (\numval{PermCtxHopThree}{+0.061}, $p$~=~\numval{PermCtxHopThreeP}{0.286}), and a negative effect at four hops (\numval{PermCtxHopFour}{\ensuremath{-}0.120}, $p$~=~\numval{PermCtxHopFourP}{0.006}). At four hops, a model free to use its own knowledge scores F1~=~\numval{PermNoCtxHopFourLevel}{0.545} without a graph and \numval{PermCorrectHopFourLevel}{0.425} with one.

That comparison is not dispositive for two reasons. Its baseline uses a third prompt: neither the strict nor the permissive template, but a plain instruction to ``answer to the best of your knowledge.'' The arms are therefore unmatched and reproduce the design defect examined in this section. The four-hop result also covers only \numval{NumQHopFour}{12} questions, of which \numval{CovFullHopFour}{4} contain their answer. The retraction applies to the original comparison; it does not establish that context can never hurt. Under a permissive prompt, the question remains open and the observed sign is negative. Resolving it requires a permissive no-context arm that has not been run.

\section{Wrong Context and the Poisoned Retrieval Path}
\label{sec:override}

The precision sweep retains the answer chain. This experiment removes it by replacing each question's correct subgraph with the subgraph of another question, selected through a deterministic offset in the question list. The substitution is tested under both prompt regimes.

Under the strict prompt, wrong-context F1 is \numval{OvStrictWrong}{0.005}, compared with \numval{OvStrictCorrect}{0.579} for the correct subgraph across \numval{OverrideStrictTrials}{2,250} trials and every model. Under the permissive prompt, wrong-context F1 rises to \numval{OvPermWrong}{0.231} (95\% CI [\numval{OvPermWrongCILo}{0.177}, \numval{OvPermWrongCIHi}{0.274}]), compared with \numval{OvPermCorrect}{0.627} for correct context across \numval{OverridePermTrials}{1,500} trials and every model. Table~\ref{tab:override} summarizes the prompt regimes, and Table~\ref{tab:override_hop} reports the strict condition by hop count. \finding{WrongContextFallsToFloor}{Under the strict prompt, wrong-context performance falls to the no-context baseline}, which is itself near zero.

\begin{table}[t]
\centering
\caption{Context override by prompt regime (strict: \numval{OverrideStrictTrials}{2,250} trials; permissive: \numval{OverridePermTrials}{1,500} trials; both use all 6 models). All values are answer F1. The no-context row is the matched-prompt baseline of Section~\ref{sec:suppression}. 95\% bootstrap CIs in brackets.}
\label{tab:override}
\begin{tabular}{lcc}
\toprule
\tblhead Condition & Strict prompt F1 & Permissive prompt F1 \\
\midrule
\tblband Correct context & \numval{OvStrictCorrect}{0.579} {\scriptsize[\numval{OvStrictCorrectCILo}{0.510},\numval{OvStrictCorrectCIHi}{0.658}]} & \numval{OvPermCorrect}{0.627} {\scriptsize[\numval{OvPermCorrectCILo}{0.561},\numval{OvPermCorrectCIHi}{0.702}]} \\
Wrong context   & \numval{OvStrictWrong}{0.005} {\scriptsize[\numval{OvStrictWrongCILo}{0.000},\numval{OvStrictWrongCIHi}{0.012}]} & \numval{OvPermWrong}{0.231} {\scriptsize[\numval{OvPermWrongCILo}{0.177},\numval{OvPermWrongCIHi}{0.274}]} \\
\tblband No context      & \numval{OvStrictNoCtxMatched}{0.035} & not run \\
\bottomrule
\end{tabular}
\end{table}

\begin{table}[t]
\centering
\caption{Wrong-context override by hop count, in answer F1 (strict prompt, all 6 models, prose). The no-context row is the matched-prompt baseline. Correct-context differs from Table~\ref{tab:matched} because this experiment is prose-only and runs its own trials.}
\label{tab:override_hop}
\begin{tabular}{lcccc}
\toprule
\tblhead Condition & 1-hop & 2-hop & 3-hop & 4-hop \\
\midrule
\tblband No context       & \numval{OvHopMatchedOne}{0.014} & \numval{OvHopMatchedTwo}{0.046} & \numval{OvHopMatchedThree}{0.044} & \numval{OvHopMatchedFour}{0.061} \\
Wrong context    & \numval{OvHopWrongOne}{0.000} & \numval{OvHopWrongTwo}{0.017} & \numval{OvHopWrongThree}{0.000} & \numval{OvHopWrongFour}{0.000} \\
\tblband Correct context  & \numval{OvHopCorrectOne}{0.875} & \numval{OvHopCorrectTwo}{0.559} & \numval{OvHopCorrectThree}{0.324} & \numval{OvHopCorrectFour}{0.215} \\
\bottomrule
\end{tabular}
\end{table}

\subsection{The comparison that would isolate suppression, and the one we have}

Comparing wrong context with \emph{correct} context combines the loss of useful facts with any suppression of the model's existing knowledge. Only the suppression component is override, so the relative drop of \numval{OvPermDropPct}{63}\% between those arms does not measure it. The contrast that would isolate it compares wrong context with no context, holding the prompt fixed across both.

We do not have that contrast. Our no-context arm uses a third prompt, neither strict nor permissive but a plain instruction to answer to the best of the model's knowledge, while the wrong-context arm uses the permissive template that invites the model to combine supplied information with what it knows. The two arms therefore differ in prompt as well as in context, which is the same design this paper retracts a finding over in \S\ref{sec:artifact}. The measured difference is \numval{OvVsNoCtx}{\ensuremath{-}0.068} F1 (95\% CI [\numval{OvVsNoCtxCILo}{\ensuremath{-}0.098}, \numval{OvVsNoCtxCIHi}{\ensuremath{-}0.039}], $p$~\numval{OvVsNoCtxP}{0.000}) over \numval{OvVsNoCtxNq}{125} questions, and the prompt change could account for part or all of it. Disabling the scorer's fallback path leaves the sign and significance unchanged, which speaks to the parser and not to the prompt mismatch.

What the number bounds is the size of any suppression effect, not its value: whatever override contributes, it is small beside the \numval{OvPermDropPct}{63}\% drop against correct context, which is the comparison that matters for a pipeline. Settling the value needs a permissive no-context arm, which we have not run.

\finding{OverrideIsModelDependent}{Whatever that difference contains, it is not uniform across models.} It inherits the prompt mismatch above, so the per-model figures are not clean suppression estimates either, but the disagreement in \emph{sign} is harder to attribute to a prompt that is identical for every model. \numval{OvModelsNoOverride}{2} of the six score \emph{higher} with a wrong subgraph than with no subgraph: GPT-5 moves from \numval{OvNoCtxGptfive}{0.334}~$\rightarrow$~\numval{OvWrongGptfive}{0.387}, and GPT-5 Mini moves from \numval{OvNoCtxMini}{0.282}~$\rightarrow$~\numval{OvWrongMini}{0.356}. These changes have the opposite sign from override. Haiku scores \numval{OvHaikuWrong}{0.027} with a wrong subgraph, while Haiku and Gemini Flash lose almost everything. Because the models disagree even on the sign, the evidence does not support treating override as an intrinsic property of structured context.

\subsection{What this does and does not say about a poisoned graph}
\label{sec:security}

\finding{PoisonedGraphIsWorseThanItLooks}{Treating the \numval{OvPermWrong}{0.231} score as resilience to a poisoned graph, or as a performance floor, would understate the exposure.} Under the permissive prompt, wrong-context F1 is \numval{NovWrongKnew}{0.551} when the model could already answer without a graph and \numval{NovWrongNothing}{0.006} when it could not. The surviving performance comes from parametric recall despite the substituted context. On a graph the model has not memorised, that source of recall is unavailable by construction.

\textbf{Substitution is not a poisoning experiment.} Three properties limit the supported conclusion. First, the condition simultaneously removes the question's evidence and adds unrelated evidence, so their effects cannot be separated. Second, the substituted triples come from an unrelated question and were not selected by an adversary or chosen to conflict semantically with the correct answer. Third, the strict condition also instructs the model to rely on the supplied facts. The experiment establishes that tolerance of benign distractors does not imply robustness to wholesale context substitution. It does not measure an attack.

\textbf{What follows for the retrieval path.} \finding{GuardTheWritePath}{The write path into a knowledge base carries risks that are already guarded against on the prompt path.} Models supplied with wrong triples do not signal that the context is wrong, and this failure requires no adversarial text in the prompt. Because the experiment included no adversary, we treat this as a property worth controlling for and do not claim a measured vulnerability. Establishing severity would require triples selected to conflict with the correct answer, an attacker model specifying who can write to the knowledge base, and a condition that substitutes evidence without also removing it. Prior work has begun this analysis for text corpora~\cite{zou2024poisonedrag} and for individual adversarial triples~\cite{zhao2025ragsafety}; this study does not extend those experiments.

The difference across prompt regimes should not be interpreted as a severity gradient. Wrong-context F1 is \numval{OvStrictWrong}{0.005} under the strict prompt and \numval{OvPermWrong}{0.231} under the permissive prompt, but neither level measures resilience. The strict score reflects a model declining to answer, while the permissive score reflects answers drawn from memory. The between-prompt comparison also crosses the parser boundary described in \S\ref{sec:unmeasurable}. We therefore rely on the within-prompt contrasts, not on the difference between these levels.\section{Three Nulls, and Four Smaller Levers}
\label{sec:nulls}

Serialization, triple ordering, and subgraph size have no measurable effect at multi-hop depth. Together with retrieval precision from Section~\ref{sec:precision}, they constitute the four nulls in Contribution 4. Every comparison in this section holds the prompt fixed across arms and is therefore unaffected by the scorer problem in Section~\ref{sec:unmeasurable}.

\subsection{Serialization affects parsing, and the active ingredient is the predicate name}
\label{sec:format}

\begin{figure}[htbp]
\centering
\includegraphics[width=\columnwidth]{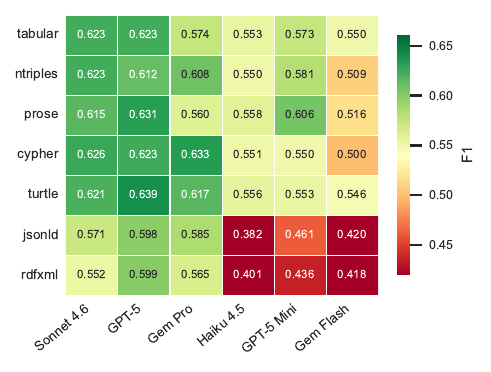}
\caption{F1 by format and model in Phase 1. The two-tier pattern appears across all models. The three mid-tier models in the right columns are more sensitive to format, consistent with a capability equalizer effect.}
\label{fig:heatmap}
\end{figure}

Phase 1 crossed all seven formats with six models and \numval{NumQuestions}{125} questions in entity-centric triple order, for \numval{PhaseOneTrials}{5,250} trials. The formats form two tiers, visible as horizontal bands in Fig.~\ref{fig:heatmap}. The top five---prose, tabular, Turtle, N-Triples, and Cypher---cluster between F1~=~\numval{FmtTopTierLow}{0.580} and \numval{FmtTopTierHigh}{0.589}. JSON-LD (\numval{FmtJsonldAvg}{0.503}) and RDF/XML (\numval{FmtRdfxmlAvg}{0.495}) trail them by \numval{FmtLagLow}{0.08}--\numval{FmtLagHigh}{0.09}.

\begin{figure}[htbp]
\centering
\includegraphics[width=\columnwidth]{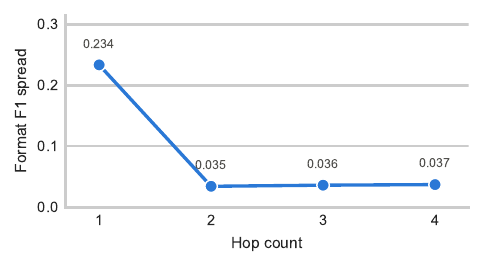}
\caption{Format F1 spread, measured as max $-$ min across seven formats, by hop count. The spread is concentrated at 1-hop, collapses once multi-hop reasoning is required, and remains flat thereafter.}
\label{fig:spread}
\end{figure}

This separation is confined to parsing and disappears with reasoning depth. The F1 spread across formats is \numval{SpreadHopOne}{0.234} at 1-hop and \numval{SpreadHopThree}{0.036} at 3-hop (Figure~\ref{fig:spread}). Once a question requires multi-hop chaining, the formats are not distinguishable on this evidence. The pooled spread is \numval{SpreadPooled}{0.093}, but that quantity is the maximum minus the minimum across seven groups and consequently overstates the likely value of choosing a format.

\textbf{Predicate vocabulary is aliased with syntax.} Verbosity does not explain the tier split. N-Triples averages \numval{TokensNtriples}{2,383} tokens per subgraph, compared with \numval{TokensProse}{644} for prose. It is by far the most verbose format and remains in the top tier, while the more compact RDF/XML and JSON-LD formats occupy the bottom tier. The split instead aligns exactly with predicate naming. The \numval{FmtWithPredLabel}{5} serializers that emit human-readable predicate labels do so through prose, table columns, Cypher relationship types, or N-Triples comments. The \numval{FmtWithoutPredLabel}{2} serializers that emit only Wikidata P-numbers are JSON-LD and RDF/XML, precisely the two bottom-tier formats. Among the \numval{FmtWithPredLabel}{5} label-bearing formats, the 1-hop spread is \numval{SpreadHopOneLabelled}{0.021}, compared with \numval{SpreadHopOne}{0.234} across all seven.

The design cannot identify a vocabulary effect independently of syntax because the two are aliased across conditions. It establishes that models perform poorly when required to infer that \texttt{wdt:P166} means ``award received''. That effect is real, practically useful, and plausibly accounts for the entire 1-hop spread, but it is not an isolated effect of serialization syntax. Separating the two requires a $2\times2$ comparison of syntax and predicate labelling, including labelled JSON-LD and RDF/XML alongside prose and Cypher with bare P-numbers. That experiment was not run. \finding{SerializationIsParsingOnly}{The format results therefore apply only to the serializers as implemented here.}

Figure~\ref{fig:heatmap} shows the complete format-by-model comparison. The format spread is \numval{SpreadMidtier}{0.160} for mid-tier models and \numval{SpreadFrontier}{0.062} for frontier models, a ratio of \numval{SpreadRatio}{2.6}. This greater sensitivity among weaker models is consistent with the capability equalizer effect reported in code generation~\cite{canedo2026equalizer}. Here, however, the sensitivity is limited to 1-hop; the two tiers converge at 3-hop.

Average evidence faithfulness across all format and model pairs is \numval{FaithOverall}{0.85}. Prose reaches \numval{FaithProse}{0.902}, tabular reaches \numval{FaithTabular}{0.907}, and RDF/XML falls to \numval{FaithRdfxml}{0.798}. Its hallucination rate is \numval{HallucRdfxml}{0.202}, compared with \numval{HallucProse}{0.098} for prose. This ordering supports a parsing account for the markup formats, although answer accuracy and citation faithfulness remain distinct. Turtle has the \emph{highest} answer F1 among the seven formats (\numval{FmtTurtleAvg}{0.589}) and the lowest faithfulness (\numval{FaithTurtle}{0.796}). A format can therefore support correct answers while weakening the citations offered to justify them.

The initial Cypher faithfulness result reflected a scoring defect, not a model property. Cypher scored \numval{FaithCypherBefore}{0.686}, initially the lowest of the seven, because models cited node variables and \textsc{screaming\_snake} relationship types from the input while the evaluator matched label triples. Correct citations expressed in Cypher's own notation consequently received zero credit. The defect affected \numval{CypherRescoredTrials}{144} trials and no other format. Resolving the variables before matching raises Cypher to \numval{FaithCypher}{0.839}. Turtle's deficit remains after the same check and is therefore treated as a real effect.

\subsection{Triple ordering}

Path-centric ordering places the chain triples consecutively, while random ordering scatters them through the prompt. Across \numval{LinTrials}{3,000} trials---two orders $\times$ two formats $\times$ six models $\times$ \numval{NumQuestions}{125} questions---the overall $\Delta$F1 is \numval{LinDelta}{+0.006}. The cleaner comparison is prose, which preserves the requested input order; there, the difference is \numval{LinDeltaProse}{\ensuremath{-}0.003}. \finding{OrderingHasNoEffect}{Triple ordering has no measurable effect.} A follow-up using \numval{LinSizeTrials}{3,000} trials and two hundred triples, padded with cross-question distractors, also finds no interaction with scale ($\Delta_{200}$~=~\numval{LinDeltaSizeTwoHundred}{\ensuremath{-}0.005}, $\Delta_{50}$~=~\numval{LinDeltaSizeFifty}{+0.009}). \finding{OrderingNullAtScale}{The ordering null also holds at two hundred triples.}

\subsection{Subgraph size, and a selection artifact}
\label{sec:distractors}

Subgraph retrieval commonly treats irrelevant triples as noise that should be filtered~\cite{dai2024llms,mavromatis2024gnnrag}. The unmatched comparison appeared to contradict that assumption, but the apparent effect came from question selection.

Across \numval{SizeTrials}{1,842} trials and six models, subgraph size varied among three conditions: chain-only, containing just the gold reasoning path; twenty-five triples; and \numval{SizeFiftyTarget}{50} triples. Chain-only scored \numval{SizeChainUnmatched}{0.591}, below the twenty-five-triple condition at \numval{SizeTwentyFiveUnmatched}{0.737} and the fifty-triple condition at \numval{SizeFiftyUnmatched}{0.649}. Faithfulness also appeared to rise from \numval{SizeChainUnmatchedFaith}{0.888} to \numval{SizeTwentyFiveUnmatchedFaith}{0.950}. Taken at face value, these values imply that pruning to the reasoning chain is harmful, a result not reported in prior work.

\textbf{The arms contained different questions.} The runner skips a size condition whenever a question's subgraph is already below the target, because padding would alter the intended comparison. That rule is reasonable within a question but invalidates an unmatched comparison across arms. The conditions ran on \numval{SizeChainNQ}{125} / \numval{SizeTwentyFiveNQ}{77} / \numval{SizeFiftyNQ}{105} questions. Because subgraph size correlates with reasoning depth, the resulting question sets also differ in difficulty. The twenty-five-triple arm contains no question above 3 hops and averages \numval{SizeTwentyFiveMeanHops}{1.49} hops, compared with \numval{SizeChainMeanHops}{2.11} for chain-only. \finding{SizeIsSelectionArtifact}{The apparent size effect was caused by an easier question set.}

\begin{figure}[htbp]
\centering
\includegraphics[width=\columnwidth]{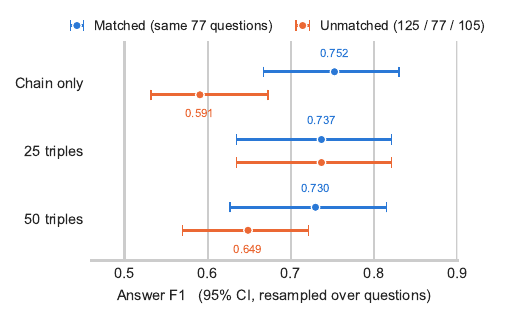}
\caption{Subgraph size has no measurable effect after matching question sets. Orange shows each arm on the questions for which it ran, reproducing the selection artifact. Blue restricts every arm to the \numval{SizeCommonQ}{77} shared questions; the resulting column is flat and every interval overlaps. The twenty-five-triple arm already defines the common set, so its two rows contain the same trials.}
\label{fig:distractors}
\end{figure}

Restricting the analysis to the \numval{SizeCommonQ}{77} questions shared by all three arms reverses the ordering and removes the effect (Table~\ref{tab:distractors}; Fig.~\ref{fig:distractors}). With trials paired by question and bootstrap resampling over questions, every contrast crosses zero on both evaluation axes. Moving from chain-only to twenty-five triples changes F1 by \numval{SizeDeltaFI}{\ensuremath{-}0.016} (95\% CI [\numval{SizeDeltaFICILo}{-0.075}, \numval{SizeDeltaFICIHi}{0.029}]) and faithfulness by \numval{SizeDeltaFaith}{+0.016} ([\numval{SizeDeltaFaithCILo}{-0.006}, \numval{SizeDeltaFaithCIHi}{0.037}]). Excluding the two OpenAI models, whose truncation rates vary across arms, produces an almost flat sequence of \numval{SizeChainMatchedNoOpenAI}{0.728} / \numval{SizeTwentyFiveMatchedNoOpenAI}{0.727} / \numval{SizeFiftyMatchedNoOpenAI}{0.725}. The larger experiment, matched by construction across all \numval{NumQuestions}{125} questions, agrees: 50~$\rightarrow$~200 triples changes F1 by \numval{SizeTwoHundredDelta}{+0.007} ([\numval{SizeTwoHundredCILo}{-0.006}, \numval{SizeTwoHundredCIHi}{0.018}]).

\begin{table}[t]
\centering
\caption{Subgraph size before and after matching question sets (\numval{SizeTrials}{1,842} trials, six models). The unmatched comparison reflects question selection. Among the \numval{SizeCommonQ}{77} questions present in every arm, no contrast is distinguishable from zero. The $n_q$ column gives the number of questions scored in each arm. F1 and faithfulness are trial means; brackets give 95\% bootstrap CIs over questions.}
\label{tab:distractors}
\begin{tabular}{lcccc}
\toprule
\tblhead & \multicolumn{2}{c}{Unmatched} & \multicolumn{2}{c}{Matched (\numval{SizeCommonQ}{77}q)} \\
\tblhead \cmidrule(lr){2-3}\cmidrule(lr){4-5}
\tblhead Condition & $n_q$ & F1 & F1 & Faith. \\
\midrule
\tblband Chain-only & \numval{SizeChainNQ}{125} & \numval{SizeChainUnmatched}{0.591} {\scriptsize[\numval{SizeChainUnCILo}{0.532},\numval{SizeChainUnCIHi}{0.672}]} & \numval{SizeChainMatched}{0.752} {\scriptsize[\numval{SizeChainMaCILo}{0.667},\numval{SizeChainMaCIHi}{0.830}]} & \numval{SizeChainMatchedFaith}{0.935} \\
25 triples & \phantom{0}\numval{SizeTwentyFiveNQ}{77} & \numval{SizeTwentyFiveUnmatched}{0.737} {\scriptsize[\numval{SizeTwentyFiveUnCILo}{0.635},\numval{SizeTwentyFiveUnCIHi}{0.821}]} & \numval{SizeTwentyFiveMatched}{0.737} {\scriptsize[\numval{SizeTwentyFiveMaCILo}{0.635},\numval{SizeTwentyFiveMaCIHi}{0.821}]} & \numval{SizeTwentyFiveMatchedFaith}{0.950} \\
\tblband 50 triples & \numval{SizeFiftyNQ}{105} & \numval{SizeFiftyUnmatched}{0.649} {\scriptsize[\numval{SizeFiftyUnCILo}{0.570},\numval{SizeFiftyUnCIHi}{0.721}]} & \numval{SizeFiftyMatched}{0.730} {\scriptsize[\numval{SizeFiftyMaCILo}{0.627},\numval{SizeFiftyMaCIHi}{0.815}]} & \numval{SizeFiftyMatchedFaith}{0.954} \\
\bottomrule
\end{tabular}
\end{table}

The range is worth stating exactly, because it is narrower than a summary suggests. The smallest arm is chain-only, which retains the full gold path and removes the distractors around it; it is not an empty prompt, and the strict no-context results elsewhere show an empty prompt behaves nothing like it. The largest is two hundred triples, and it comes from a separate experiment over a different question set with cross-question padding. There is no paired chain-only against two hundred contrast. What the two experiments jointly support is that with the answer path retained, adding distractors up to fifty changed nothing on the shared questions, and a further expansion from fifty to two hundred changed nothing either. Within that range, these results neither confirm nor refute the assumption that irrelevant triples are noise. They do show that subgraph size is not a measurable lever at this scale, consistent with both the ordering null above and the precision null in Section~\ref{sec:precision}.

\subsection{Four further levers}

\textbf{Evidence-first prompting (\numval{EvidenceFirstTrials}{1,500} trials, six models).} Requiring the evidence block \emph{before} the answer increased faithfulness by \numval{EvidenceFirstFaith}{+0.028} and reduced F1 by \numval{EvidenceFirstFI}{\ensuremath{-}0.021}. The models identified relevant triples when required to cite them first, but better evidence selection did not produce better answers. \finding{GroundingIsNotReasoning}{Improved grounding did not improve reasoning.}

\textbf{Chain-of-thought prompting (\numval{CotTrials}{1,500} trials, six models).} Adding ``think step by step, follow connections one hop at a time'' to the system prompt reduced performance at every hop count above 1-hop: $\Delta$F1~=~\numval{CotDeltaHopTwo}{\ensuremath{-}0.020} at 2-hop, \numval{CotDeltaHopThree}{\ensuremath{-}0.043} at 3-hop, and \numval{CotDeltaHopFour}{\ensuremath{-}0.031} at 4-hop. \finding{CotHurts}{Step-by-step instructions compounded errors and did not aid traversal.} A two-model pilot had suggested a gain at 2-hop, but that pattern did not replicate across all six models.

\textbf{Dynamic oracle scoping (\numval{DynamicTrials}{2,250} trials, six models).} Each multi-hop SPARQL query was decomposed into per-hop steps, and each step received only its oracle-correct triples plus the preceding step's answer. Dynamic oracle scored F1~=~\numval{DynOracle}{0.520}, below both full-context at \numval{DynFullCtx}{0.578} and chain-only at \numval{DynChainOnly}{0.604}. \finding{ScopingLosesHolisticView}{Externally imposed decomposition removed the holistic view needed by the model.} StructGPT~\cite{jiang2023structgpt} instead allows the LLM to control which triples it explores, distinguishing model-driven exploration from the externally imposed decomposition tested here.

\textbf{Sampling.} A verification run at temperature~=~0.7 with five samples per trial (\numval{TempTrials}{3,750} trials across all \numval{NumQuestions}{125} questions) reproduces the main pattern. Context helps at 1-hop ($\Delta$~=~\numval{TempDeltaHopOne}{+0.726}), and wrong-context override persists (F1~=~\numval{TempWrongCtx}{0.015}). \finding{GreedyResultsHold}{Sampling preserves the condition rankings observed under greedy decoding.}

Table~\ref{tab:summary} in \S\ref{sec:intro} reports every experiment in the study and its verdict.

\section{A Contrast We Cannot Measure}
\label{sec:unmeasurable}

All preceding comparisons either hold the prompt fixed across arms or vary it while leaving both contexts empty. The remaining contrast changes the prompt while correct context is present. It asks the question most relevant to practice: what does the grounding instruction cost when retrieval succeeds?

The experiment was run, but its estimate is an artifact of the scorer. The source of that artifact prevents a defensible point estimate from being reported.

\subsection{One cause}

The strict prompt requires a structured response beginning with ``Answer: \ldots''. The permissive prompt imposes no response structure, and models generally answer it in prose. The answer parser has a primary path for the structured form and a fallback for free text. The primary path always fires in the strict arm. It often fails in the permissive arm, where the fallback then rescues the trial.

The fallback rescues \numval{ParserRescued}{76} trials, but those rescues are distributed entirely on one side of the comparison. It adds \numval{ParserGainStrict}{+0.000} to the strict arm and \numval{ParserGainPermissive}{+0.085} to the permissive arm. No strict arm in the study contains a rescued trial because the strict prompt requires the format expected by the primary parser. Every rescue occurs in a permissive arm with a subgraph in the prompt. Across the full corpus, the fallback changes the score by \numval{ParserGainOverall}{+0.003}.

The independent variable therefore determines the response format, and the scorer responds to that format. Enabling the fallback credits permissive responses that the primary parser cannot read. Disabling it penalises the permissive arm for format instead of content. Neither version isolates the effect of the grounding instruction on accuracy.

\subsection{Four symptoms}

Table~\ref{tab:signflip} distinguishes results that cross the strict/permissive prompt boundary with context present from results that do not. Both columns are computed from the same stored trials.

\begin{table}[t]
\centering
\caption{Dependence on the answer parser's fallback path. Every result in the upper block compares a strict arm with a permissive arm while correct context is present, causing the arms to produce different response formats. Results in the lower block do not cross that boundary with context present, so the parser choice does not act on them. The wrong-versus-no-context row is the exception worth naming: its arms differ in prompt as well as in context (\S\ref{sec:override}), which is a separate problem from the parser and is not repaired by the right-hand column. Values are answer F1. The right column recomputes the same trials using only the structured parse; permutation $p$ appears in parentheses. Both columns are computed from the same stored trials.}
\label{tab:signflip}
\begin{tabular}{lcc}
\toprule
\tblhead Result & As reported & Primary parse only \\
\midrule
\tblband Grounding cost, correct ctx & \numval{NovAll}{+0.045} ({\scriptsize\numval{NovAllP}{0.005}}) & \numval{NovAllPrimary}{\ensuremath{-}0.039} ({\scriptsize\numval{NovAllPrimaryP}{0.041}}) \\
Cost where it knew nothing & \numval{NovNothing}{\ensuremath{-}0.012} ({\scriptsize\numval{NovNothingP}{0.521}}) & \numval{NovNothingPrimary}{\ensuremath{-}0.090} ({\scriptsize\numval{NovNothingPrimaryP}{0.000}}) \\
\tblband Cost where it knew a little & \numval{NovLittle}{+0.042} ({\scriptsize\numval{NovLittleP}{0.126}}) & \numval{NovLittlePrimary}{\ensuremath{-}0.034} ({\scriptsize\numval{NovLittlePrimaryP}{0.298}}) \\
Cost where it knew it & \numval{NovKnew}{+0.127} ({\scriptsize\numval{NovKnewP}{0.006}}) & \numval{NovKnewPrimary}{+0.024} ({\scriptsize\numval{NovKnewPrimaryP}{0.632}}) \\
\midrule
\tblband Suppression, no facts & \numval{PromptSwingNoCtx}{0.264} & \numval{PromptSwingNoCtxPrimary}{0.264} \\
Wrong ctx vs.\ no ctx & \numval{OvVsNoCtx}{\ensuremath{-}0.068} & \numval{OvVsNoCtxPrimary}{\ensuremath{-}0.073} \\
\tblband Format spread, pooled & \numval{SpreadPooled}{0.093} & \numval{SpreadPooledPrimary}{0.093} \\
Precision at constant volume & \numval{PrecDelta}{+0.003} & \numval{PrecDeltaPrimary}{+0.003} \\
\bottomrule
\end{tabular}
\end{table}

The upper block measures one quantity in four ways. With the fallback enabled, the overall grounding cost is \numval{NovAll}{+0.045} (95\% CI [\numval{NovAllCILo}{+0.007}, \numval{NovAllCIHi}{+0.075}], $p$~=~\numval{NovAllP}{0.005}, \numval{NovAllNq}{125} questions), which rejects the null. With the fallback disabled, the same trials produce \numval{NovAllPrimary}{\ensuremath{-}0.039} (95\% CI [\numval{NovAllPrimaryCILo}{\ensuremath{-}0.079}, \numval{NovAllPrimaryCIHi}{\ensuremath{-}0.007}], $p$~=~\numval{NovAllPrimaryP}{0.041}), which also rejects the null but has the opposite sign. Opposite significant estimates from the same trials, differing only in response parsing, cannot be resolved by collecting more data.

The next three rows divide the same contrast according to what each model could answer without a graph, using the permissive no-context arm as a per-question measure of parametric knowledge. With the fallback, the pooled values form a monotone gradient from \numval{NovNothing}{\ensuremath{-}0.012} where the model knew nothing to \numval{NovKnew}{+0.127} where it knew the answer. That pattern suggests that the instruction withholds parametric fallback and therefore costs accuracy only when such a fallback exists. The parser comparison does not support that mechanism. Under the primary parse, the bands become \numval{NovNothingPrimary}{\ensuremath{-}0.090}, \numval{NovLittlePrimary}{\ensuremath{-}0.034}, and \numval{NovKnewPrimary}{+0.024}. They remain ordered but shift enough that the band previously interpreted as free rejects the null ($p$~=~\numval{NovNothingPrimaryP}{0.000}), while the band previously interpreted as expensive does not ($p$~=~\numval{NovKnewPrimaryP}{0.632}). Because the substantive interpretation reverses with the parser, the ordering alone does not identify a mechanism.

Two additional analyses use the same defective contrast. The depth-specific strict-versus-permissive handicap in Table~\ref{tab:handicap} compares strict and permissive prompts with correct context in both arms. Restricting that comparison to questions whose subgraph contains the answer does the same and yields \numval{CovHandicapHopThree}{+0.211} at 3-hop over \numval{CovHandicapHopThreeNq}{11} questions. Both are retained here only to show the shape of the parser-dependent artifact.

\begin{table}[t]
\centering
\caption{This table is not a result; it shows the artifact's shape. It compares strict and permissive prompts by hop count, with both arms using prose and correct context (\numval{PermissiveTrials}{750} trials each). Values are answer F1, and positive $\Delta$ means the strict prompt scored worse. The scorer cannot measure this contrast for the reason given in this section. Brackets contain 95\% bootstrap CIs for $\Delta$, resampling questions.}
\label{tab:handicap}
\begin{tabular}{lcccc}
\toprule
\tblhead Prompt & 1-hop & 2-hop & 3-hop & 4-hop \\
\midrule
\tblband Strict      & \numval{HandicapStrictHopOne}{0.883} & \numval{HandicapStrictHopTwo}{0.556} & \numval{HandicapStrictHopThree}{0.321} & \numval{HandicapStrictHopFour}{0.226} \\
Permissive  & \numval{HandicapPermHopOne}{0.871} & \numval{HandicapPermHopTwo}{0.541} & \numval{HandicapPermHopThree}{0.455} & \numval{HandicapPermHopFour}{0.425} \\
\tblband $\Delta$    & \numval{HandicapDeltaHopOne}{\ensuremath{-}0.012} & \numval{HandicapDeltaHopTwo}{\ensuremath{-}0.015} & \numval{HandicapDeltaHopThree}{+0.133} & \numval{HandicapDeltaHopFour}{+0.199} \\
{\scriptsize 95\% CI} & {\scriptsize[\numval{HandicapDeltaHopOneCILo}{-0.030},\numval{HandicapDeltaHopOneCIHi}{0.003}]} & {\scriptsize[\numval{HandicapDeltaHopTwoCILo}{-0.069},\numval{HandicapDeltaHopTwoCIHi}{0.037}]} & {\scriptsize[\numval{HandicapDeltaHopThreeCILo}{0.051},\numval{HandicapDeltaHopThreeCIHi}{0.216}]} & {\scriptsize[\numval{HandicapDeltaHopFourCILo}{0.078},\numval{HandicapDeltaHopFourCIHi}{0.343}]} \\
\tblband {\scriptsize perm.\ $p$} & {\scriptsize \numval{HandicapPermPHopOne}{0.127}} & {\scriptsize \numval{HandicapPermPHopTwo}{0.609}} & {\scriptsize \numval{HandicapPermPHopThree}{0.004}} & {\scriptsize \numval{HandicapPermPHopFour}{0.020}} \\
\bottomrule
\end{tabular}
\end{table}

\subsection{Two problems that are not the parser}

The covered-set restriction has another defect that would remain even after repairing the parser. Its coverage filter checks whether the gold answer string appears anywhere in the subgraph, not whether the answer is reachable from the question entity along the gold chain. Of the \numval{CovHandicapHopThreeNq}{11} questions in that subset, \numval{ChainUnbackedHopThree}{3} fail the stricter reachability test. Those same questions produce the three largest strict-prompt costs in the set. The direction follows from the filter: when an answer is present but unreachable, the strict prompt cannot reach it from the supplied facts, while the permissive prompt can use parametric knowledge. The filter therefore selects questions that penalise the strict arm. Among the \numval{ChainBackedHopThreeNq}{8} questions that pass the reachability test, the contrast is \numval{ChainBackedHopThree}{+0.114}, with an interval spanning zero (95\% CI [\numval{ChainBackedHopThreeCILo}{\ensuremath{-}0.022}, \numval{ChainBackedHopThreeCIHi}{+0.271}], $p$~=~\numval{ChainBackedHopThreeP}{0.250}). At 4-hop, the covered subset contains \numval{CovHandicapHopFourNq}{4} questions at $p$~=~\numval{CovHandicapHopFourP}{0.250} and supports no conclusion.

The strict prompt also bundles two interventions. It restricts the model to the supplied facts and fixes the response format. The evidence-first result shows that response format alone changes F1. Even a format-insensitive scorer would therefore estimate the combined effect of these instructions, not the cost of the grounding sentence by itself.

\subsection{What we would need}

A valid comparison requires three changes, none of which was tested here. First, the scorer must be validated on held-out responses to read both output formats equally well. Second, the permissive prompt must request the same structured output as the strict prompt, leaving the grounding restriction as the only difference between arms. Third, a $2\times2$ comparison of the restriction and the evidence block must separate their effects. Until those conditions are met, Table~\ref{tab:signflip} exhausts what this apparatus supports. A point estimate in either direction measures the parser, not the grounding instruction.

\subsection{A note on the corpus split}

One part of the gradient remains descriptive at the corpus level, although it does not support a mechanism. The two source corpora behave differently on this contrast. Within QALD, the strict-permissive gap is large and rejects the null, ranging from \numval{NovQaLittle}{+0.091} ($p$~=~\numval{NovQaLittleP}{0.038}) to \numval{NovQaKnew}{+0.306} ($p$~=~\numval{NovQaKnewP}{0.001}). Within LC-QuAD, the gap is absent in every band: \numval{NovLcNothing}{\ensuremath{-}0.017}, \numval{NovLcLittle}{\ensuremath{-}0.011}, and \numval{NovLcKnew}{\ensuremath{-}0.007}, with every interval spanning zero. Table~\ref{tab:novelty} reports the split.

LC-QuAD scores \numval{LcCtxFone}{0.710} with context, compared with \numval{QaCtxFone}{0.270} for QALD, so the data cannot distinguish a genuine absence of the contrast from a ceiling effect. The parser problem also applies to both corpus columns. The supported conclusion is limited but consequential: the pooled column combines two populations, and even a valid measurement of this contrast would not yield a single value for the corpus.

\begin{table}[t]
\centering
\caption{This table is not a result; it displays the artifact by source corpus and by what the model could answer without a graph. Values are differences in answer F1, with positive values indicating that the strict prompt scored worse. Bands are defined by the permissive no-context score averaged per question, and parenthesised values are question counts. Every cell depends on the parser choice described in this section.}
\label{tab:novelty}
\begin{tabular}{lccc}
\toprule
\tblhead Unaided score {\scriptsize (not a result)} & LC-QuAD & QALD & Pooled \\
\midrule
\tblband F1~=~0 & \numval{NovLcNothing}{\ensuremath{-}0.017} ({\scriptsize\numval{NovLcNothingNq}{42}}) & \numval{NovQaNothing}{+0.028} ({\scriptsize\numval{NovQaNothingNq}{6}}) & \numval{NovNothing}{\ensuremath{-}0.012} \\
$0 <$ F1 $\leq$ \numval{NovBandEdge}{0.5} & \numval{NovLcLittle}{\ensuremath{-}0.011} ({\scriptsize\numval{NovLcLittleNq}{20}}) & \numval{NovQaLittle}{+0.091} ({\scriptsize\numval{NovQaLittleNq}{22}}) & \numval{NovLittle}{+0.042} \\
\tblband F1~$>$ \numval{NovBandEdge}{0.5} & \numval{NovLcKnew}{\ensuremath{-}0.007} ({\scriptsize\numval{NovLcKnewNq}{20}}) & \numval{NovQaKnew}{+0.306} ({\scriptsize\numval{NovQaKnewNq}{15}}) & \numval{NovKnew}{+0.127} \\
\midrule
All questions & \numval{LcCtxFone}{0.710} ctx F1 & \numval{QaCtxFone}{0.270} ctx F1 & \numval{NovAll}{+0.045} \\
\bottomrule
\end{tabular}
\end{table}

\label{sec:pipeline}

Five implications follow, ordered by their expected importance to a pipeline. Each rests on the section cited with it.

\begin{itemize}
\item \textbf{\finding{SpendBudgetOnRecall}{Changing the provenance of triples around the answer has no measured benefit, while losing the answer path has a large cost.}} With the chain present, replacing every surrounding triple with material from an unrelated entity changes F1 by \numval{PrecDelta}{+0.003}. The corresponding changes are \numval{PrecCovered}{+0.004} on questions whose subgraph demonstrably contains the answer and \numval{PrecNovNothing}{+0.015} on questions the model could not answer unaided. Removing the chain reduces F1 to \numval{OvPermWrong}{0.231} under a permissive prompt and \numval{OvStrictWrong}{0.005} under a strict prompt. Sections~\ref{sec:precision} and~\ref{sec:override} report these two results.

The evidence for composition is more complete than the evidence for recall. Composition was varied from \numval{PrecCtxPrecZero}{1.00} context precision to \numval{PrecCtxPrecHundred}{0.51}, with a flat result across the gradient. Recall was observed only at the endpoints because the wrong-context condition removes all evidence for the question. The study therefore establishes a large difference between a complete answer path and no answer path, but it does not measure intermediate recall levels.

Pipeline settings are one step removed from these measurements. The experiments varied the number of triples reaching the prompt and the fraction belonging to the question; they did not build an index, run similarity search, or test a retriever. The supported interpretation is that retriever output with the measured properties loses nothing, not that any particular retrieval setting has been validated. Raising $k$ and declining to prune correspond most directly to the tested conditions. Widening the hop radius and lowering a similarity threshold are inferential extensions because they are useful only if they exchange precision for recall along the same axis. Within the tested range, that exchange costs tokens but no measured accuracy, up to two hundred triples and a context that is half wrong.

This asymmetry supports biasing retrieval toward recall instead of tuning for observable precision. Production recall is unavailable as a direct signal because identifying the triples on the answer chain requires solving the question. Optimizing the metric that can be observed would therefore optimize the term that had no measured effect here.

\item \textbf{\finding{GuardTheWritePath}{Guard the write path into the knowledge base.}} Models do not detect a wrong subgraph, so the retrieval path needs controls comparable to those applied to the prompt path. Relative to no context, average suppression is \numval{OvVsNoCtx}{\ensuremath{-}0.068} F1, and \numval{OvModelsNoOverride}{2} of the six models show none. The effect is therefore smaller and less uniform than the raw correct-versus-wrong comparison implies. The aggregate also masks the greater exposure on an unmemorised graph, where wrong-context F1 is \numval{NovWrongNothing}{0.006}.

\item \textbf{Match the prompt across arms in every evaluation you run.} This is the least expensive control in the paper, and omitting it produced an apparently publishable but artifactual finding. If the context arm says ``use only the provided facts'', the no-context arm must carry the same instruction. Section~\ref{sec:suppression} measures the consequence of that difference. Section~\ref{sec:unmeasurable} adds a second requirement: before comparing prompt regimes, verify that the scorer reads both response formats equally well.

\item \textbf{\finding{NamePredicates}{Include human-readable names for predicates.}} The apparent one-hop format gap tracks whether a serializer emits the predicate label or only its identifier. The spread among the \numval{FmtWithPredLabel}{5} formats that include labels is \numval{SpreadHopOneLabelled}{0.021}, compared with \numval{SpreadHopOne}{0.234} across all seven. Because syntax and vocabulary are aliased in this design, the actionable intervention is to add labels; the evidence does not support changing syntax. A label lookup is also cheaper than replacing a serializer.

\item \textbf{\finding{StopTuning}{Stop tuning four choices that produced no measurable benefit.}} Subgraph size is flat from chain-only to two hundred triples, triple ordering is flat at every tested scale, serialization has no measurable effect at multi-hop depth, and retrieval precision is flat at constant volume. \finding{GiveWholeSubgraph}{Give the model the complete subgraph}: per-hop scoping scores \numval{DynOracle}{0.520} even when every step receives oracle-correct triples, compared with \numval{DynFullCtx}{0.578} for full context. The evidence provides no support for withholding context and some evidence that doing so is harmful.
\end{itemize}

The study does not support a recommendation about the grounding instruction itself. Section~\ref{sec:unmeasurable} explains why. The instruction is close to free when the model has no parametric answer to suppress. When retrieval returns nothing, it separates \numval{MatchedNoCtxOverall}{0.035} from \numval{PermNoCtxOverall}{0.299}. Its cost when retrieval succeeds cannot be measured by this apparatus.

\section{Limitations}
\label{sec:limitations}

\textbf{The oracle subgraphs do not guarantee oracle coverage.} They are called oracle subgraphs because they are constructed from gold SPARQL instead of retrieved. That construction does not ensure that the extracted neighbourhood retains the answer. Coverage was checked by testing whether every gold answer label appeared anywhere in each question's subgraph. Full coverage declines with depth: \numval{CovFullHopOne}{46} of \numval{CovAnswerableHopOne}{46} at one-hop, \numval{CovFullHopTwo}{30} of \numval{CovAnswerableHopTwo}{36} at two-hop, \numval{CovFullHopThree}{11} of \numval{CovAnswerableHopThree}{23} at three-hop, and \numval{CovFullHopFour}{4} of \numval{CovAnswerableHopFour}{11} at four-hop. Another \numval{CovNone}{7} subgraphs contain none of their gold answers. The totals exclude \numval{CovAggregate}{7} aggregate questions whose answers are counts and therefore could not appear in a subgraph.

Because depth and benchmark are aliased, coverage is more plainly described as a corpus property: \numval{LcCovered}{76} of \numval{LcQuestions}{82} for LC-QuAD and \numval{QaCovered}{16} of \numval{QaQuestions}{43} for QALD. The depth curve reflects those corpus-level values through the hop split. These cases are extraction failures, not evidence that the questions themselves are difficult. For example, the four-hop subgraph for ``Where is the poet Alexander Pope buried?'' contains triples about Alexander von Humboldt, Jenson Button and Alex Salmond because entity linking selected the wrong Alexander. The subgraphs were not repaired because all trials used them as extracted.

Hop count is the number of predicates in the SPARQL string, not verified sequential chain depth. Every question at three or more hops was therefore classified from the structure of its gold SPARQL. The classifier expands property paths and repeated-subject shorthand. It gives priority to transitive paths, classifies three or more predicates on one node as fan-out, assigns queries with a residual filter to mixed, and treats the remainder as chains. Table~\ref{tab:sparql} reports all \numval{NumQClassified}{43} of the \numval{NumQHighHop}{43} questions in this group. The structural classes are unevenly represented, which limits comparisons among them. Context benefit is positive for every class: \numval{SparqlBenefitSeqChain}{+0.284} for sequential chains, \numval{SparqlBenefitTransitive}{+0.242} for transitive closures, \numval{SparqlBenefitFanout}{+0.170} for fan-outs and \numval{SparqlBenefitMixed}{+0.129} for mixed patterns. However, only \numval{SparqlTypesRejecting}{2} of \numval{SparqlTypesTotal}{4} reject at \numval{AlphaLevel}{0.05}, and the smallest class contains \numval{SparqlBenefitNqMixed}{5} questions. \finding{StructureVariesWithinDepth}{No structural class points in the opposite direction}, but that does not establish that the results hold across classes.

\begin{table}[t]
\centering
\caption{SPARQL structural classification of all \numval{NumQClassified}{43} questions at three or more hops, derived from the gold queries by the classifier described above. No type contradicts the others, although two are too small to test. Structure varies within a hop count, so this proxy measures predicate count and not verified reasoning depth. Each example gives the shape found in the question set, with bound entities omitted and repeated subjects written in full. Fan-out denotes three or more predicates that constrain one node.}
\label{tab:sparql}
\begin{tabular}{lcp{4.2cm}}
\toprule
\tblhead Type & Count & Example Pattern \\
\midrule
\tblband Sequential chain & \numval{NumSparqlSeqChain}{11} & \texttt{?a P1 ?b . ?b P2 ?c} \\
Transitive closure & \numval{NumSparqlTransitive}{17} & \texttt{?x P1/P279* ?y} \\
\tblband Fan-out & \numval{NumSparqlFanout}{10} & \texttt{?x P1 ?a . ?x P2 ?b .} \newline \texttt{?x P3 ?c} \\
Mixed & \numval{NumSparqlMixed}{5} & \texttt{?x P1 ?a . ?x P2 ?b .} \newline \texttt{FILTER(...)} \\
\bottomrule
\end{tabular}
\end{table}

\textbf{Depth cannot be separated from benchmark.} All one- and two-hop questions come from LC-QuAD, while all three- and four-hop questions come from QALD. Every depth-stratified result is therefore also a between-corpus result, and this design cannot identify which difference produced it. The corpora vary simultaneously in answer-set size, coverage, difficulty and the sign of the context effect. This is the paper's most consequential limitation because depth organizes most of the reported results. Separation would require both benchmarks to supply questions at every depth, which neither does.

\textbf{The scorer is format-sensitive and was not validated across formats.} Section~\ref{sec:unmeasurable} gives the consequence for one contrast. More generally, an end-task metric computed by a parser is comparable across conditions only if the parser is equally valid for each response format. Any independent variable that changes response format therefore requires validation on both arms before comparison. This parser was validated on neither format; the defect became visible only after the fallback was disabled and the analysis rerun.

\textbf{Training-data contamination.} All \numval{NumQuestions}{125} questions concern Wikidata entities likely to have appeared in pretraining data, while graph retrieval is also deployed over graphs that have not. The permissive no-context arm measures the relevant parametric knowledge per question and is used to stratify both the precision null and the wrong-context result. It remains a proxy. An unaided score of zero establishes only that the model could not answer that question; it does not establish that the entity was absent from training. A difficult question about a famous entity falls into the same band. A genuine test requires a private graph or post-cutoff facts, neither of which was used here.

\textbf{Coverage falls with depth.} Full gold coverage holds for \numval{CovFullTotal}{92} of \numval{CovAnswerable}{117} answerable questions. The rate falls from \numval{CovPctHopOne}{100.00}\% at one-hop to \numval{CovPctHopFour}{36.36}\% at four-hop. Coverage and depth consequently move together in every depth-stratified result. The deepest questions are therefore the least reliable both because coverage is lowest and because the group is small.

\textbf{Sample size at high hops.} The four-hop group contains only \numval{NumQHopFour}{12} questions, and \numval{MatchedHopFourNoBenefitQ}{7} show no context benefit. Its findings are directionally consistent with the three-hop results but underpowered, so no significance is claimed. The percentile bootstrap excludes zero, while the exact permutation test gives $p$~=~\numval{MatchedPermPHopFour}{0.109}. No multiplicity correction is applied across the roughly forty intervals in the paper, and individual $p$ values should be interpreted accordingly.

\textbf{Oracle subgraph construction.} Chain triples are identified from gold SPARQL queries and are not retrieved. The study contains no retriever, so ``retrieval precision'' describes a property of constructed context, not a setting on an implemented system. Real retrievers may omit chain triples in ways the sweep does not represent, and experiments with genuine retrieval may produce different results.

\textbf{The precision sweep covers four models, not the full set.} The strongest practical recommendation rests on \numval{PrecTrials}{1,999} trials over \numval{PrecModels}{4} models from the six-model set, using only prose serialization and entity-centric order. The other models were not run, and the null is not claimed for them. Each condition also draws its wrong triples from a single donor question. Those triples are consequently more coherent than the mistakes of a real retriever may be.

\textbf{Plain text is the only tested channel.} Every condition supplies the subgraph as plain text in the prompt, as current GraphRAG pipelines do. The results do not cover graph context introduced during tokenization, represented by learned graph tokens, or passed through an encoder as embeddings. Because the observed format effects are parsing effects, a channel that bypasses text parsing might eliminate them or move them into the encoder. That channel remains to be tested.

\textbf{No iterative baseline.} Every condition injects a fixed subgraph. Systems in which the model chooses subsequent retrievals~\cite{sun2024tog,jiang2023structgpt} use a different and stronger multi-hop design. No such system is implemented here, so the static findings cannot be claimed to transfer to it.

\textbf{Three of the six models are moving targets.} All models were accessed through one gateway, and half of the identifiers are mutable aliases instead of pinned snapshots. The API-reported resolutions on the twenty-sixth of August, two thousand twenty-six, were \modelid{gpt-5-2025-08-07}, \modelid{gpt-5-mini-2025-08-07}, \modelid{claude-haiku-4-5-20251001}, \modelid{claude-sonnet-4-6}, \modelid{gemini-2.5-pro}, \modelid{gemini-2.5-flash}. The latter three cannot be pinned retrospectively, so exact replication of their results is not guaranteed.

\textbf{Two output-token configurations.} A total of \numval{TruncatedTrials}{971} trials from the two OpenAI models returned no visible text and were rerun with a higher token cap, leaving those models represented by two configurations. These reasoning models count reasoning tokens among reported completion tokens, and the original \numval{MaxTokensOld}{2,048} cap was exhausted before an answer appeared. The reruns used \numval{MaxTokensNew}{16,384}. Trials that had already returned text were not rerun because the cap had not bound for them and decoding was greedy. Blank responses correlate with question difficulty, so this repair is not missing-at-random and preferentially upgrades the hardest questions. Four-hop results are therefore also reported with an OpenAI-excluded sensitivity analysis.

\textbf{Hop-count measurement.} Hop count records predicate count in the SPARQL string, not verified sequential chain depth. Classification of all \numval{NumQClassified}{43} high-hop questions into chains, transitive closures, fan-outs and mixed patterns finds no class that contradicts the others. The measure nonetheless remains a proxy and reflects benchmark authors' choices about expressing relations as property paths or explicit chains.

\textbf{Two prompt regimes.} The experiments cover two regimes and one rephrased strict template. They do not characterize the grounding instructions between those conditions. The strict prompt also combines a restriction with a response format, and those components were not varied independently.

\section{Conclusion}
\label{sec:conclusion}

Across \numval{NumExperiments}{16} experiments and \numval{TotalTrials}{30,841} trials, two of the four subgraph-related choices in a GraphRAG pipeline change the answer. They occur at the beginning and end of the pipeline.

The first is whether the answer chain reaches the prompt. Once it does, changing whether the remaining context belongs to the question is worth \numval{PrecDelta}{+0.003} F1 at constant volume. That null holds at every tested depth and among questions the model could not answer unaided. Retrieval quality therefore has two components with sharply different measured value, and only recall warrants an accuracy budget. The intervening choices produce no measurable benefit at multi-hop depth. Serialization spread falls from \numval{SpreadHopOne}{0.234} at one-hop to \numval{SpreadHopThree}{0.036} at three-hop, and the one-hop gap tracks the presence of a human-readable predicate name, not syntax alone. Triple order has no effect at either tested scale, and subgraph size is flat from chain-only to two hundred triples after matching the question sets.

The second consequential choice is the grounding instruction, which is also where the measurement limitations arise. With no facts in the prompt, the instruction suppresses parametric recall by a factor of \numval{SuppressionFactor}{8.63}. This is the largest effect in the study, but it characterizes an evaluation with empty context and not a functioning retrieval pipeline. Applying the instruction to a context arm while omitting it from the baseline produces an artifactual finding that graph context hurts at depth. With prompts matched across arms, context helps at one-, two- and three-hop and remains inconclusive at four-hop.

The practically important contrast is the cost of that instruction when correct context is available, and this apparatus cannot measure it. Prompt regime determines response format, the answer scorer handles those formats differently, and changing the parser reverses the estimated effect while rejecting the null in both directions. Resolving the contrast requires a scorer validated across both formats and a permissive prompt that requests the same output structure as the strict prompt.

\bibliographystyle{IEEEtran}
\bibliography{references}

\begin{thebibliography}{10}
\providecommand{\url}[1]{#1}
\csname url@samestyle\endcsname
\providecommand{\newblock}{\relax}
\providecommand{\bibinfo}[2]{#2}
\providecommand{\BIBentrySTDinterwordspacing}{\spaceskip=0pt\relax}
\providecommand{\BIBentryALTinterwordstretchfactor}{4}
\providecommand{\BIBentryALTinterwordspacing}{\spaceskip=\fontdimen2\font plus
\BIBentryALTinterwordstretchfactor\fontdimen3\font minus
  \fontdimen4\font\relax}
\providecommand{\BIBforeignlanguage}[2]{{%
\expandafter\ifx\csname l@#1\endcsname\relax
\typeout{** WARNING: IEEEtran.bst: No hyphenation pattern has been}%
\typeout{** loaded for the language `#1'. Using the pattern for}%
\typeout{** the default language instead.}%
\else
\language=\csname l@#1\endcsname
\fi
#2}}
\providecommand{\BIBdecl}{\relax}
\BIBdecl

\bibitem{edge2024graphrag}
\BIBentryALTinterwordspacing
D.~Edge \emph{et~al.}, ``From local to global: A graph {RAG} approach to
  query-focused summarization,'' \emph{arXiv preprint arXiv:2404.16130}, 2024.
  [Online]. Available: \url{https://arxiv.org/abs/2404.16130}
\BIBentrySTDinterwordspacing

\bibitem{dai2024llms}
\BIBentryALTinterwordspacing
X.~Dai, Y.~Hua, T.~Wu, Y.~Sheng, Q.~Ji, and G.~Qi, ``Large language models can
  better understand knowledge graphs than we thought,'' \emph{arXiv preprint
  arXiv:2402.11541}, 2024. [Online]. Available:
  \url{https://arxiv.org/abs/2402.11541}
\BIBentrySTDinterwordspacing

\bibitem{mavromatis2024gnnrag}
\BIBentryALTinterwordspacing
C.~Mavromatis and G.~Karypis, ``{GNN-RAG}: Graph neural retrieval for large
  language model reasoning,'' \emph{arXiv preprint arXiv:2405.20139}, 2024.
  [Online]. Available: \url{https://arxiv.org/abs/2405.20139}
\BIBentrySTDinterwordspacing

\bibitem{fatemi2024talk}
\BIBentryALTinterwordspacing
B.~Fatemi, J.~Halcrow, and B.~Perozzi, ``Talk like a graph: Encoding graphs for
  large language models,'' in \emph{International Conference on Learning
  Representations (ICLR)}, 2024. [Online]. Available:
  \url{https://arxiv.org/abs/2310.04560}
\BIBentrySTDinterwordspacing

\bibitem{sui2024table}
\BIBentryALTinterwordspacing
Y.~Sui, M.~Zhou, M.~Zhou, S.~Han, and D.~Zhang, ``Table meets {LLM}: Can large
  language models understand structured table data? a benchmark and empirical
  study,'' in \emph{ACM International Conference on Web Search and Data Mining
  (WSDM)}, 2024. [Online]. Available: \url{https://arxiv.org/abs/2305.13062}
\BIBentrySTDinterwordspacing

\bibitem{frey2023turtle}
\BIBentryALTinterwordspacing
J.~Frey \emph{et~al.}, ``Benchmarking the abilities of large language models
  for {RDF} knowledge graph creation and comprehension: How well do {LLMs}
  speak {Turtle}?'' in \emph{DL4KG Workshop at ISWC}, 2023. [Online].
  Available: \url{https://arxiv.org/abs/2309.17122}
\BIBentrySTDinterwordspacing

\bibitem{perozzi2024letgraph}
\BIBentryALTinterwordspacing
B.~Perozzi, B.~Fatemi \emph{et~al.}, ``Let your graph do the talking: Encoding
  structured data for {LLMs},'' \emph{arXiv preprint arXiv:2402.05862}, 2024.
  [Online]. Available: \url{https://arxiv.org/abs/2402.05862}
\BIBentrySTDinterwordspacing

\bibitem{wu2024clasheval}
\BIBentryALTinterwordspacing
K.~Wu, E.~Wu, and J.~Zou, ``{ClashEval}: Quantifying the tug-of-war between an
  {LLM}'s internal prior and external evidence,'' in \emph{Advances in Neural
  Information Processing Systems (NeurIPS), Datasets and Benchmarks Track},
  2024. [Online]. Available: \url{https://arxiv.org/abs/2404.10198}
\BIBentrySTDinterwordspacing

\bibitem{xie2024chameleon}
\BIBentryALTinterwordspacing
J.~Xie, K.~Zhang, J.~Chen, R.~Lou, and Y.~Su, ``Adaptive chameleon or stubborn
  sloth: Revealing the behavior of large language models in knowledge
  conflicts,'' in \emph{International Conference on Learning Representations
  (ICLR), Spotlight}, 2024. [Online]. Available:
  \url{https://arxiv.org/abs/2305.13300}
\BIBentrySTDinterwordspacing

\bibitem{zou2024poisonedrag}
\BIBentryALTinterwordspacing
W.~Zou, R.~Geng, B.~Wang, and J.~Jia, ``{PoisonedRAG}: Knowledge corruption
  attacks to retrieval-augmented generation of large language models,'' in
  \emph{USENIX Security Symposium}, 2025. [Online]. Available:
  \url{https://arxiv.org/abs/2402.07867}
\BIBentrySTDinterwordspacing

\bibitem{zhao2025ragsafety}
\BIBentryALTinterwordspacing
T.~Zhao \emph{et~al.}, ``{RAG} safety: Exploring knowledge poisoning attacks to
  retrieval-augmented generation,'' \emph{arXiv preprint arXiv:2507.08862},
  2025. [Online]. Available: \url{https://arxiv.org/abs/2507.08862}
\BIBentrySTDinterwordspacing

\bibitem{zhou2023context}
\BIBentryALTinterwordspacing
W.~Zhou, S.~Zhang, H.~Poon, and M.~Chen, ``Context-faithful prompting for large
  language models,'' in \emph{Findings of the Association for Computational
  Linguistics: EMNLP}, 2023. [Online]. Available:
  \url{https://arxiv.org/abs/2303.11315}
\BIBentrySTDinterwordspacing

\bibitem{huang2024trust}
\BIBentryALTinterwordspacing
Y.~Huang, S.~Chen, H.~Cai, and B.~Dhingra, ``To trust or not to trust?
  enhancing large language models' situated faithfulness to external
  contexts,'' in \emph{International Conference on Learning Representations
  (ICLR)}, 2025. [Online]. Available: \url{https://arxiv.org/abs/2410.14675}
\BIBentrySTDinterwordspacing

\bibitem{bi2024context}
\BIBentryALTinterwordspacing
B.~Bi, S.~Huang, Y.~Wang \emph{et~al.}, ``Context-{DPO}: Aligning language
  models for context-faithfulness,'' \emph{arXiv preprint arXiv:2412.15280},
  2024. [Online]. Available: \url{https://arxiv.org/abs/2412.15280}
\BIBentrySTDinterwordspacing

\bibitem{bi2025params}
\BIBentryALTinterwordspacing
B.~Bi, S.~Liu, Y.~Wang, Y.~Xu, L.~Mei, J.~Fang, and X.~Cheng, ``Parameters vs.\
  context: Fine-grained control of knowledge reliance in language models,''
  \emph{arXiv preprint arXiv:2503.15888}, 2025. [Online]. Available:
  \url{https://arxiv.org/abs/2503.15888}
\BIBentrySTDinterwordspacing

\bibitem{huang2025parammute}
\BIBentryALTinterwordspacing
P.~Huang, Z.~Liu, Y.~Yan \emph{et~al.}, ``{ParamMute}: Suppressing
  knowledge-critical {FFNs} for faithful retrieval-augmented generation,'' in
  \emph{Advances in Neural Information Processing Systems (NeurIPS)}, 2025.
  [Online]. Available: \url{https://arxiv.org/abs/2502.15543}
\BIBentrySTDinterwordspacing

\bibitem{mandarapu2026grounding}
\BIBentryALTinterwordspacing
M.~Mandarapu and S.~Kunkunuru, ``Knowledge-graph grounding helps {LLMs} only
  for out-of-training knowledge: A controlled study on clinical question
  answering,'' \emph{arXiv preprint arXiv:2606.22419}, 2026. [Online].
  Available: \url{https://arxiv.org/abs/2606.22419}
\BIBentrySTDinterwordspacing

\bibitem{sun2024tog}
\BIBentryALTinterwordspacing
J.~Sun, C.~Xu, L.~Tang, S.~Wang, C.~Lin, Y.~Gong, L.~M. Ni, H.-Y. Shum, and
  J.~Guo, ``Think-on-graph: Deep and responsible reasoning of large language
  model on knowledge graph,'' in \emph{International Conference on Learning
  Representations (ICLR)}, 2024. [Online]. Available:
  \url{https://arxiv.org/abs/2307.07697}
\BIBentrySTDinterwordspacing

\bibitem{jiang2023structgpt}
\BIBentryALTinterwordspacing
J.~Jiang, K.~Zhou, Z.~Dong, K.~Ye, W.~X. Zhao, and J.-R. Wen, ``{StructGPT}: A
  general framework for large language model to reason over structured data,''
  in \emph{Proceedings of the Conference on Empirical Methods in Natural
  Language Processing (EMNLP)}, 2023. [Online]. Available:
  \url{https://arxiv.org/abs/2305.09645}
\BIBentrySTDinterwordspacing

\bibitem{shan2026bounded}
\BIBentryALTinterwordspacing
X.~Shan and Y.~Luo, ``Bounded path context: A controlled study of visible path
  history in {LLM}-based knowledge graph question answering,'' \emph{arXiv
  preprint arXiv:2605.26645}, 2026, code:
  https://github.com/AndyShan11/Bounded-Path-Context. [Online]. Available:
  \url{https://arxiv.org/abs/2605.26645}
\BIBentrySTDinterwordspacing

\bibitem{nguyen2024cot}
\BIBentryALTinterwordspacing
M.-V. Nguyen \emph{et~al.}, ``Direct evaluation of chain-of-thought in
  multi-hop reasoning with knowledge graphs,'' in \emph{Findings of the
  Association for Computational Linguistics: ACL}, 2024. [Online]. Available:
  \url{https://arxiv.org/abs/2402.11199}
\BIBentrySTDinterwordspacing

\bibitem{canedo2026equalizer}
\BIBentryALTinterwordspacing
A.~Canedo, ``Architecture as capability equalizer for coding agents,'' 2026.
  [Online]. Available:
  \url{https://github.com/arquicanedo/architecture-as-equalizer}
\BIBentrySTDinterwordspacing

\bibitem{vrandecic2014wikidata}
D.~Vrande\v{c}i\'{c} and M.~Kr\"{o}tzsch, ``Wikidata: A free collaborative
  knowledgebase,'' \emph{Communications of the ACM}, vol.~57, no.~10, pp.
  78--85, 2014.

\bibitem{dubey2019lcquad2}
\BIBentryALTinterwordspacing
M.~Dubey, D.~Banerjee, A.~Abdelkawi, and J.~Lehmann, ``{LC-QuAD} 2.0: A large
  dataset for complex question answering over {Wikidata} and {DBpedia},'' in
  \emph{International Semantic Web Conference (ISWC)}, 2019. [Online].
  Available: \url{https://jens-lehmann.org/files/2019/iswc_lcquad2.pdf}
\BIBentrySTDinterwordspacing

\bibitem{usbeck2023qald10}
R.~Usbeck \emph{et~al.}, ``{QALD}-10 --- the 10th challenge on question
  answering over linked data,'' in \emph{Semantic Web Challenges (SemWebEval)},
  2023.

\bibitem{w3c2014turtle}
\BIBentryALTinterwordspacing
D.~Beckett, T.~Berners-Lee, E.~Prud'hommeaux, and G.~Carothers, ``{RDF} 1.1
  {Turtle},'' W3C, Recommendation, 2014. [Online]. Available:
  \url{https://www.w3.org/TR/turtle/}
\BIBentrySTDinterwordspacing

\bibitem{w3c2014ntriples}
\BIBentryALTinterwordspacing
G.~Carothers and A.~Seaborne, ``{RDF} 1.1 {N-Triples},'' W3C, Recommendation,
  2014. [Online]. Available: \url{https://www.w3.org/TR/n-triples/}
\BIBentrySTDinterwordspacing

\bibitem{w3c2014jsonld}
\BIBentryALTinterwordspacing
M.~Sporny, G.~Kellogg, and M.~Lanthaler, ``{JSON-LD} 1.0,'' W3C,
  Recommendation, 2014. [Online]. Available:
  \url{https://www.w3.org/TR/json-ld/}
\BIBentrySTDinterwordspacing

\bibitem{w3c2004rdfxml}
\BIBentryALTinterwordspacing
D.~Beckett, ``{RDF/XML} syntax specification (revised),'' W3C, Recommendation,
  2004. [Online]. Available: \url{https://www.w3.org/TR/rdf-syntax-grammar/}
\BIBentrySTDinterwordspacing

\end{thebibliography}

\end{document}